\documentclass[10pt,journal,compsoc]{IEEEtran}

\ifCLASSOPTIONcompsoc
\usepackage[nocompress]{cite}
\else
\usepackage{cite}
\fi

\usepackage{bm}
\usepackage{mathrsfs}
\usepackage{times}
\usepackage{epsfig}
\usepackage{graphicx}
\usepackage{subfigure}
\usepackage{amsmath}
\usepackage{amssymb}
\usepackage{color}
\usepackage[noend]{algpseudocode}
\usepackage{algorithmicx,algorithm}
\ifCLASSINFOpdf
\else
\fi

\begin{document}
	
	\title{RobustFusion: Robust Volumetric Performance Reconstruction under Human-object Interactions from Monocular RGBD Stream}
	
	%
	
	\author{Zhuo Su$^*$, Lan Xu$^*$, Dawei Zhong$^*$, Zhong Li, Fan Deng, Shuxue Quan and Lu FANG$^{\S}$
		\IEEEcompsocitemizethanks{
			\IEEEcompsocthanksitem $*:$ Equal Contribution
			\IEEEcompsocthanksitem 
			Z. Su, D. Zhong and L. Fang are with Dept. of Electronic Engineering at Tsinghua University, and Beijing National Research Center for Information Science and Technology. 
			Z. Su is also with Dept. of Automation, Tsinghua University. 
			Lan Xu is with School of Information Science and Technology, Shanghaitech University. Fan Deng, Zhong Li and Shuxue Quan are with OPPO US Research Center.
				\IEEEcompsocthanksitem This work is supported in part by Natural Science Foundation of China (NSFC) under contract No. 61860206003 and 62088102, in part by OPPO.   
			\IEEEcompsocthanksitem ${\S}$ Correspondence Author: \protect E-mail: fanglu@tsinghua.edu.cn}
	}
	
	\markboth{}%
	{Shell \MakeLowercase{\textit{et al.}}: Bare Demo of IEEEtran.cls for Computer Society Journals}
	
	\IEEEtitleabstractindextext{%
		\begin{abstract}
			High-quality 4D reconstruction of human performance with complex interactions to various objects is essential in real-world scenarios, which enables numerous immersive VR/AR applications.
			However, recent advances still fail to provide reliable performance reconstruction, suffering from challenging interaction patterns and severe occlusions, especially for the monocular setting.
			To fill this gap, in this paper, we propose RobustFusion, a robust volumetric performance reconstruction system for human-object interaction scenarios using only a single RGBD sensor, which combines various data-driven visual and interaction cues to handle the complex interaction patterns and severe occlusions.
			We propose a semantic-aware scene decoupling scheme to model the occlusions explicitly, with a segmentation refinement and robust object tracking to prevent disentanglement uncertainty and maintain temporal consistency.
			We further introduce a robust performance capture scheme with the aid of various data-driven cues, which not only enables re-initialization ability, but also models the complex human-object interaction patterns in a data-driven manner.
			To this end, we introduce a spatial relation prior to prevent implausible intersections, as well as data-driven interaction cues to maintain natural motions, especially for those regions under severe human-object occlusions.
			We also adopt an adaptive fusion scheme for temporally coherent human-object reconstruction with occlusion analysis and human parsing cue.
			Extensive experiments demonstrate the effectiveness of our approach to achieve high-quality 4D human performance reconstruction under complex human-object interactions whilst still maintaining the lightweight monocular setting.
		\end{abstract}
		
		}

	\maketitle

	\IEEEdisplaynontitleabstractindextext

	%
	\IEEEpeerreviewmaketitle

	\IEEEraisesectionheading{\section{Introduction}\label{sec:introduction}}
	\label{Introduction}
	
\IEEEPARstart{T}{he} 
rise of virtual reality and augmented reality (VR and AR) to present information in an innovative and immersive way has increased the demand for human-centric 4D (3D spatial plus 1D temporal) content generation, with various applications from entertainment to commerce, from gaming to education, from military to art.
Further, reconstruct the 4D models of human activities under human-object interactions both robustly and conveniently remains unsolved, which suffers from challenging interaction patterns and severe occlusions.
It evolves as a cutting-edge yet bottleneck technique and has recently attracted substantive attention of both the computer vision and computer graphics communities.

Early model-based methods~\cite{ATevs2012amimation, NJ2007Dynamic, li2009robust, HaoliTemplate, Templaterealtime} suffer from pre-scanned templates or inefficient run-time performance, which are unacceptable for daily interactive application.
Recent volumetric approaches have eliminated the reliance on the templates and increased both the effectiveness and efficiency with modern GPUs.
\begin{figure}[tbp]
	\centering
	\subfigure[RGBD stream (Input)]{\includegraphics[height=110pt]{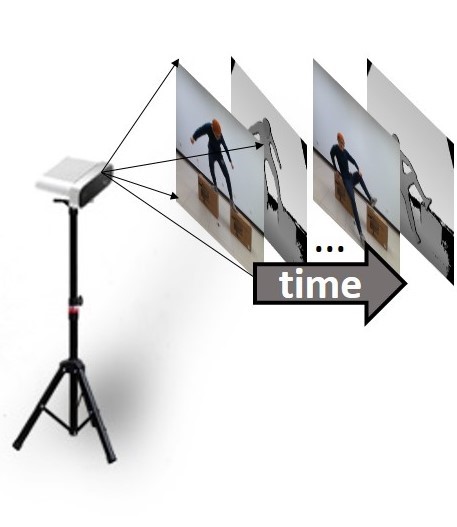}}
	\subfigure[4D reconstruction (Output)]{\includegraphics[height=110pt]{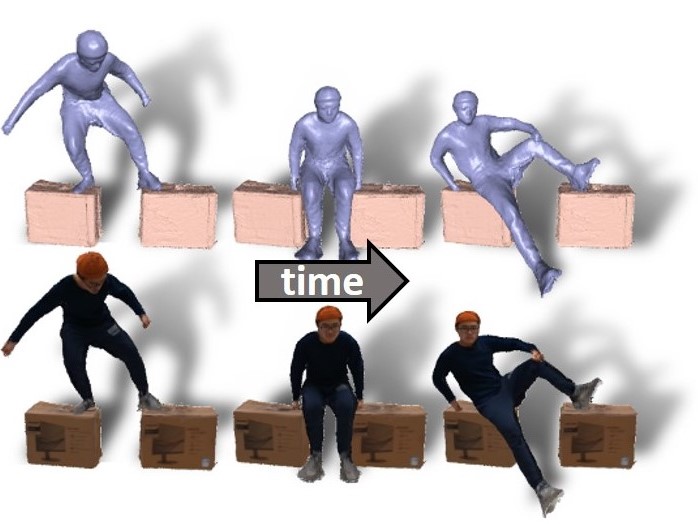}}	   
	\caption{Illustration of the system and results of our RobustFusion}
	\label{fig:teaser}
\end{figure}
The high-end solutions~\cite{collet2015high,dou-siggraph2016,motion2fusion,TotalCapture,UnstructureLan} achieve realistic human-object reconstruction using multi-view studio setup which provides sufficient view observation to solve the challenging interaction and occlusion ambiguity.
However, their complex and expensive multi-view studio setting leads to the high restriction of the wide daily applications.	
Differently, the monocular volumetric approaches adopt the handiest commercial RGBD camera and temporal fusion pipeline.
Early general solutions~\cite{newcombe2015dynamicfusion,innmann2016volume,KillingFusion2017cvpr,mixedfusion2018} handle general dynamic scenes without disentangling human and objects, suffering from careful and orchestrated motions.
Recent solutions~\cite{BodyFusion, DoubleFusion, li2021posefusion} embed the human parametric models into the fusion pipeline to handle more complex motions.
Within this category, our conference version RobustFusion~\cite{RobustFusion2020ECCV} (denoted as RobustFusion(Conf.)) further enables more robust monocular capture using various data-driven visual cues such as motion~\cite{OpenPose,HMR18}, geometry~\cite{PIFU_2019ICCV,DeepHuman_2019ICCV} or semantic 
segmentation~\cite{BodyPix}.
It gets rid of the self-scanning constraint for monocular capture with re-initialization ability, where the captured performer does not need to turn around carefully to obtain complete reconstruction.
However, these monocular approaches with human priors neglect to model the mutual influence between human and object, leading to limited reconstruction under the challenging interaction scenarios.	
%
On the other hand, various researchers~\cite{PSI2019,zhang2020object,2020phosa_Arrangements,PLACE:3DV:2020,HPS,Hassan:CVPR:2021,PatelCVPR2021,GRAB:2020} reconstruct the 4D relations between humans and the objects or the environments.
However, they only recover the naked human bodies or heavily rely on specific pre-scanned object and scene templates to optimize the spatial arrangement.
Researchers pay less attention to strengthen the template-less volumetric performance capture by utilizing the rich human-object interaction priors, especially for the monocular setting.

%
In this paper, we attack the above challenges and propose \textit{RobustFusion}, a robust human-object volumetric performance capture system combined with various data-driven visual and interaction cues using only a single RGBD sensor (with optional multi-view setup).
As illustrated in Fig.~\ref{fig:teaser}, our approach solves the challenging ambiguity and severe occlusions under complex human-object interactions, achieving robust volumetric performance reconstruction, which outperforms the baselines favorably without using any pre-scanned templates.

Combining data-driven cues for robust volumetric reconstruction under challenging human-object interactions is non-trivial, let alone maintaining the lightweight property and fast running performance under the monocular setting.
To encode the interaction pattern and alleviate the occlusion ambiguity, our key idea is to utilize the data-driven interaction cues for human motions prior under occlusions, as well as the rich visual cues including scene semantic segmentation, body part parsing estimation, implicit occupancy learning, and human pose and shape detection.	
%
More specifically, we first embrace the scene semantic cue for scene decoupling to model the challenging occlusions explicitly for human-object interactions.
To prevent the disentanglement uncertainty, we refine the human-object segmentation through robust object tracking in an iterative manner, which utilizes previous reconstruction results for temporal consistency.
We also adopt a human initialization in the first frame similar to RobustFusion(Conf.), which utilizes the human parsing and implicit occupancy learning to generate a complete and fine-detailed initial model and non-rigid motion for the human.
Such initialization eliminates the tedious self-scanning constraint for more robust human-object performance capture. 
Then, we propose a robust human performance capture scheme with the aid of various data-driven cues.
Besides the original strategy with the human pose, shape, and parsing priors to enable re-initialization ability similar to RobustFusion(Conf.), we further model the interaction patterns for the challenging human-object occlusions in a data-driven manner.
To this end, we introduce a novel spatial relation prior to prevent physically implausible intersections, as well as the interaction poses prior based on Gauss Mixture Model (GMM) and the temporal interaction prior based on LSTM predictor to maintain natural motions, especially for those regions under severe occlusions.
%
Finally, we adopt an adaptive fusion scheme to obtain temporally coherent reconstruction results.
With both the human-object occlusion analysis and human parsing cue, the fusion weights are adaptively adjusted to avoid deteriorated fusion caused by tracking errors and occlusions. 
To summarize, our main contributions include:
\begin{itemize} 
	\setlength\itemsep{0em}
	\item We propose a robust volumetric performance reconstruction approach for challenging human-object interaction scenarios using only a single RGBD camera, which embraces data-driven visual and interaction cues to achieve significant superiority to existing state-of-the-art methods.

	\item We introduce a novel scene decoupling scheme under the volumetric capture framework for explicit disentanglement of human-object interactions, with the aid of robust object tracking and semantic refinement.
	
	\item We propose a novel and robust human-object performance capture scheme with various data-driven interaction cues, which can handle challenging human motions with complex interaction patterns and severe occlusions.

\end{itemize}

	\section{Related Work}\label{sec:Literature}
	\label{Literature}
	\graphicspath{{./Related Works/}}
This section presents an overview of research works related to the proposed RobustFusion system. We roughly divide the related methods into human volumetric capture, object-related reconstruction, and data-driven human and scene visual cues.

\begin{figure*}[tbp]
	\centering
	\includegraphics[width=0.99\textwidth]{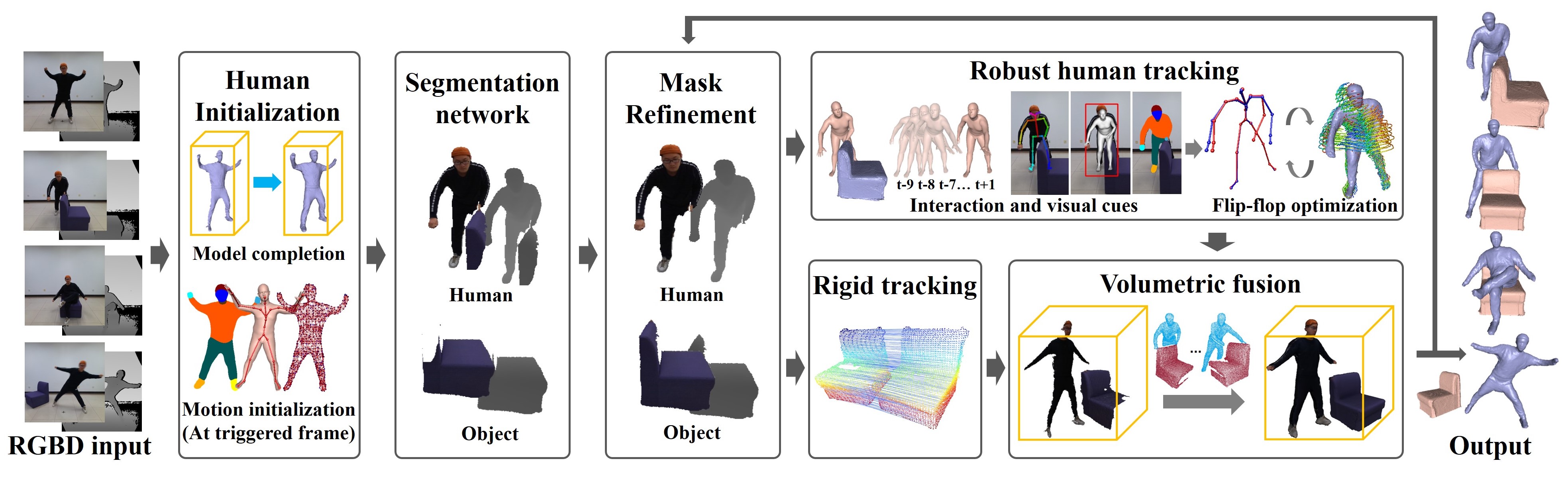}
	\caption{The pipeline of RobustFusion. Assuming monocular RGBD input, our approach consists of a human initialization stage only at the triggered frame (Sec.~\ref{Sec:initialization}), a scene decoupling stage that includes mask refinement and object tracking (Sec.~\ref{Sec: Object}), a robust human tracking stage(Sec.~\ref{Sec:human_track}) and volumertic fusion stage (Sec.~\ref{Sec: TSDF_fusion}) to generate live 4D results.}
	\label{fig_overview}
    \end{figure*}

\noindent\textbf{Human Volumetric Capture.}
	In recent years, free-form dynamic reconstruction methods combine the volumetric fusion~\cite{Curless1996} and the non-rigid tracking based on embedded deformation~\cite{sumner2007embedded}. 
	The multi-view solutions~\cite{dou-siggraph2016,motion2fusion} rely on complex studio and are difficult to be deployed for daily usage. Although \cite{UnstructureLan} achieves an unstructured multi-view setup but still not as convenient as the single-view setting.
	In contrast, \cite{newcombe2015dynamicfusion} utilizes only one common single RGBD camera and achieves real-time dynamic reconstruction.
	Then, the solution~\cite{innmann2016volume} adds the SIFT features to improve the accuracy of motion reconstruction, while Guo~\emph{et al.}~\cite{guo2017real} utilize shading information to improve the non-rigid registration and estimate the surface appearance,
	and Slavcheva~\emph{et al.}~\cite{slavcheva2018cvpr,slavcheva2020pami} add more constraints to support topology changes.
	Yu~\emph{et al.}~\cite{BodyFusion,DoubleFusion,DoubleFusionDetection} further take human articulated skeleton prior into account to increase tracking robustness, 
	while HybridFusion~\cite{HybridFusion} utilizes extra IMU sensors for more reliable motion tracking and Xu~\emph{et al.}~\cite{FlyFusion} model the mutual gains between capture view selection and reconstruction.
	Besides, POSEFusion~\cite{li2021posefusion} combines both implicit inference network and temporal volumetric fusion in a keyframe selection scheme and can capture more dynamic details in invisible regions, and some methods~\cite{SemiPara_2019CVPR,LookinGood} combine the neural rendering techniques.
	Above methods still suffer from careful and orchestrated motions, especially for a tedious self-scanning process where the performer needs to turn around carefully to obtain complete reconstruction, and RobustFusion(Conf.)~\cite{RobustFusion2020ECCV} liberates this constraint by introducing implicit occupancy method~\cite{PIFU_2019ICCV}. However, these methods either cannot handle modeling human-object interactions (e.g.,\cite{BodyFusion,DoubleFusion,UnstructureLan,HybridFusion,li2021posefusion,RobustFusion2020ECCV}) or cannot robustly handle the fast human non-rigid motions (e.g.,~\cite{KillingFusion2017cvpr, NJ2007Dynamic}).
	Comparably, our approach is more robust for capturing challenging motions under human-object interaction scenarios with the re-initialization ability and enables the simultaneous reconstruction of both human and object without the self-scanning constraint.

	\noindent\textbf{Object-related Reconstruction.}
    As for object reconstruction, the method~\cite{huang2012occlusion} utilizes structure-from-motion to recover complete 3D models of articulated objects and analyses its joint movement from RGB images. \cite{park2017colored} introduces color information to point cloud registration, which makes the tracking and reconstruction of rigid objects more robust from RGBD data. \cite{liu2018object} uses an object-aware guidance approach for autonomous scene scanning and reconstruction. Apart from rigid objects, the reconstruction of non-rigid objects explored in work ~\cite{newcombe2015dynamicfusion} mentioned above tracks the dynamically growing node points and then fuses to a canonical model. Besides, \cite{mixedfusion2018} reconstruct dynamic objects and static indoor environment at the same time. Note that the dynamic motions that \cite{newcombe2015dynamicfusion} and \cite{mixedfusion2018} capture are very limited.
    In addition to the traditional reconstruction methods for objects, recovering the 3D shape of an object from single or multiple RGB images using deep neural networks has attracted increasing attention in the past few years. \cite{Park_2019_CVPR_Deepsdf} proposes a learned continuous Signed Distance Function representation of the object. \cite{sitzmann2019scene} encodes both the geometry and appearance of the objects and represents objects as continuous functions that map world coordinates to a feature representation of local properties. Besides, \cite{xie2020pix2vox++} uses a well-designed encoder-decoder to generate a coarse 3D volume from each input image, and then a multi-scale context-aware fusion module is introduced to adaptively select high-quality reconstructions for different parts from all coarse 3D volumes to obtain a fused 3D volume, in which the network is also widely used to handle human-object interaction.  
    Moreover, \cite{wei2013modeling,wei2016modeling} propose a 4D human-object interaction model to detect human-object geometric relation and the interaction events. 
    Recently, the work~\cite{2020phosa_Arrangements} learns the spatial arrangements of humans and objects with pose estimation in a 3D scene from a single RGB image. The methods~\cite{PSI2019,PLACE:3DV:2020,Hassan:CVPR:2021} try to generate the plausible human model(s) in existing 3D scenes, and the work~\cite{HPS} utilizes wearable sensors to estimate the human pose and location in a 3D scene. However, they are limited to the naked human body or the pre-scanned models. 
    For tiny objects, the work~\cite{GRAB:2020} also provides a dataset of whole-body human grasping of objects, and the method~\cite{GrapingField:3DV:2020} proposes an expressive representation for human grasp modeling.
    By investigating and exploiting these object-related reconstruction methods, we propose our object tracking and reconstruction scheme in the meantime of our robust human volumetric capture.
    
	\noindent\textbf{Data-driven Human and Scene Visual Cues.}
    For data-driven human modeling problems, early human modeling methods~\cite{Shotton2011,Ganapathi10} take advantage of data-driven machine learning strategies to convert the capture problem into regression or pose classification problems.
    Recently, data-driven techniques have attracted more and more interest due to the rise of deep learning and RGB-based human modeling approaches bloom.
	First, the methods~\cite{OpenPose,Rogez16,Mehta2017} estimate human 2D or 3D skeletal pose, and human parametric models~\cite{SCAPE2005,SMPL2015} with human pose and shape parameters provide a good sparse representation for human models, based on which some recent work~\cite{HMR18,Kovalenko2019arXiv,joo2020eft,keepitSMPL, zhang2020object} also learn the human shape.
	Note that the use of Gauss Mixture Model from\cite{keepitSMPL} provides us an idea of utilizing a data-driven pose distribution cue for human-object interaction scenarios. Besides, LSTM-based temporal pose prediction is also taken into account.
	Second, there are many approaches that directly estimate the human geometry from RGB images, such as the parametric representation~\cite{alldieck2019tex2shape, alldieck19cvpr}, implicit representation~\cite{PIFU_2019ICCV,saito2020pifuhd,huang2020arch,zheng2020pamir} and volume representation~\cite{varol18_bodynet,DeepHuman_2019ICCV}.
	However, such predicted geometry lacks fine details, which is important for immersive human modeling. Even if ~\cite{saito2020pifuhd} achieves finer geometry details, its learned pose is also as inaccurate as the above methods.
	Besides, some performance capture methods~\cite{MonoPerfCap,LiveCap2019tog,eventCap2020CVPR,deepcap} based on RGB videos leverage the above learnable pose detection~\cite{OpenPose,Mehta2017} or its own pose regression network to improve the accuracy of human motion capture, but these methods have to rely on pre-scanned template models.
	As for scene visual cues,
	scene segmentation methods~\cite{zhao2017pspnet,he2017mask} fetch the semantic information of the whole scene including the people and objects in it,
	and human parsing methods~\cite{IntoPerson_CVPR2017,HumanParsing_ECCV2018,BodyPix} also propose to fetch the semantic information of the human model.
	These data-driven methods yield colossal potential for human performance reconstruction. We explore building a robust human and object volumetric capture algorithm on top of these priors and then achieve significant superiority to previous methods.

	\section{ Overview}\label{sec:Overview}
	\label{Overview}

    RobustFusion achieves both human and objects volumetric capture under a unified framework in a model-specific way, which can perform human-object interaction reconstructions and maintain robust ability to handle challenging human motions.
	As illustrated in Fig.~\ref{fig_overview}, our approach takes an RGBD video from Kinect v2 (or Kinect Azure) as input and generates 4D meshes, achieving more robust results than previous methods considerably.
	In our volumetric capture framework, we utilize TSDF~\cite{Curless1996} volume for geometry reconstruction, just like in~\cite{DoubleFusion,UnstructureLan}.
    A brief introduction of our technical components is provided as follows:
    
	\noindent\textbf{Human initialization.} 
	First, for model initialization, we follow~\cite{RobustFusion2020ECCV} to generate a high-quality watertight human model with fine geometric details at the beginning, in which we combine the implicit occupancy regression network with the traditional non-rigid fusion pipeline using only the front-view RGBD input.
	Second, we further utilize the complete model to initialize both the human motions and the visual priors before the tracking stage. We adopt a hybrid motion representation, including the newly sampled ED-graph and embedded SMPL. Besides, various human pose and parsing priors based on the front-view input are associated with the initialized model.
	
	\noindent\textbf{Scene Decoupling.}
	To reconstruct the dynamic scene, we first apply a semantic segmentation network to obtain the foreground masks, including both human and labeled objects. Note that human masks can also be obtained from the Kinect SDK, and as for the unlabeled objects in the network, we extract its masks by utilizing background separation. The segmented masks are too coarse to be applied for tracking. Thus, with the help of the reconstructed results, we can project 3D models to current 2D image and use a iterative strategy to refine the masks.  
	
	\noindent\textbf{Object Tracking.}
	After scene decoupling, we track the rigid motions of the objects by solving an optimization problem under the Iterative Closest Point (ICP) framework by taking account of color consistency, geometry consistency, and spatial relationship between the human and objects. The correct decoupling results provided by mask refinement enable accurate object tracking, and the correct tracked object models provide a good reference for scene decoupling in turn.
	
	\noindent\textbf{Robust Human Tracking.} 
	The core of our pipeline is to solve the hybrid motion parameters from the canonical frame to the current camera view.
	We propose a robust human tracking scheme which utilizes reliable interaction and visual data-driven priors to optimize both the skeletal and surface motions in an iterative flip-flop manner.
	Observed that human poses have particular patterns in the interaction with objects, we train a GMM model and LSTM predictor to exploit the spatial and temporal prior information in the optimization. Moreover, our scheme can handle challenging motions with the re-initialization ability.

	\noindent\textbf{Object-aware Reconstruction.} 
	We fuse the masked depth stream into the canonical TSDF volume after motion tracing to provide temporal-coherent results for the human and objects separately. The human model is fused based on the non-rigid motion field, and the object model is fused based on the estimated rigid transformation.
	Based on various visual priors and object-aware occlusion ratios, we adaptively adjust the fusion weight to avoid deteriorated fusion caused by tracking errors and occlusions.
	Finally, dynamic atlas~\cite{UnstructureLan} and per-vertex color fusion are adopted to obtain 4D textured reconstruction results.
	
	\label{Algorithm}
		
	\section{Technical Details}
	
	\subsection{Problem Representation}\label{Sec:problem}
    Motion tracking is a core problem in our performance capture system. To robustly estimate both human and object motions, we decouple and track them separately with the data-driven cues. This subsection briefly overviews these motion representations and defines the mathematical notations in our tracking framework.
    
    The motion of rigid objects is formulated by the rigid transformations ${\mathbf{T}}=\{T_{i}, i \in N \}$ in $\mathbf{SE(3)}$ space, where $N$ is the number of objects. As for human motions, we adopt the efficient and robust double-layer surface representation for motion representation~\cite{DoubleFusion}, which combines the embedded deformation (ED) and the linear human model SMPL~\cite{SMPL2015}. Since we can get a complete human model after model initialization (Sec.\ref{Sec:initialization}), we modify the SMPL-sampled ED-graph by the ED-graph sampled on the complete model.  
    We utilize SMPL to represent our skeleton motions. SMPL is a linear body model with $N =
    6890$ vertices and $K = 24$ joints. Before posing, the body model $\bar{\textbf{T}}$ deforms into the morphed model $T(\bm{\beta}, \bm{\theta})$ with the shape parameters $\bm{\beta}$ and pose parameters $\bm{\theta}$ as $T(\bm{\beta}, \bm{\theta}) = \bar{\textbf{T}}+B_s(\bm{\beta}) + B_p( \bm{\theta})$,
    where $B_s(\bm{\beta})$ and $ B_p( \bm{\theta})$ are the shape blendshapes and pose blendshapes respectively. $T(\bar{\mathbf{v}};\bm{\beta}, \bm{\theta})$ denotes the morphed 3D position for any vertex $\bar{\mathbf{v}} \in \bar{\textbf{T}}$.
    The posed SMPL is further formulated as the blend skinning function: $W(T(\bm{\beta}, \bm{\theta}), J(\bm{\beta}), \bm{\theta}, \mathcal{W})$, in terms of the body $T(\bm{\beta}, \bm{\theta})$, pose parameters $\bm{\theta}$, joint locations $J(\bm{\beta})$ and the skinning weights $\mathcal{W}$. Specifically, for any 3D vertex $\mathbf{v}_c$, the linear blend skinning (LBS) operation with the SMPL skeleton motions is formulated as $\hat{\mathbf{v}}_c =\mathbf{G}(\mathbf{v}_c,\bm{\theta})\mathbf{v}_c$,
    where $\mathbf{G}(\mathbf{v}_c,\bm{\theta})=\sum\limits_{i\in\mathcal{B}}w_{i,v_c}\mathbf{G}_i$ is the posed rigid transformation of $\mathbf{v}_c$, $\mathcal{B}$ is index set of bones, $\mathbf{G}_i = \prod\limits_{k\in\mathcal{K}_i}\mathrm{exp}(\theta_k\hat{\xi}_k)$ is the rigid transformation of $i$-th bone referencing the parent bones whose indices are $\mathcal{K}_i$ in the backward kinematic chain, $\mathrm{exp}(\theta_k\hat{\xi}_k)$ is the exponential map of the twist associated with $k$-th bone, and $w_{i,v_c}$ is the skinning weight associated with $i$-th bone and point $\mathbf{v}_c$.
    If $\mathbf{v}_c$ is on SMPL model, $w_{i,v_c}$ is pre-defined in $\mathcal{W}$. If $\mathbf{v}_c$ is on the fused surface, $w_{i,v_c}$ is given by the weighted average of its knn-nodes.
    
    Non-rigid motions of the human is represented by a embedded deformation node-graph $G=\{\mathbf{dq}_j, \mathbf{x}_j\}$, consisting of the dual quaternions $\{\mathbf{dq}_j\}$ and the corresponding ED nodes $\{\mathbf{x}_j\}$. $SE3(\mathbf{dq}_j)$ denotes the rigid transformation in $\mathbf{SE(3)}$ space. Then for any 3D vertex $\mathbf{v}_c$ in the canonical volume, the ED warping operation is formulated as follows:
    \begin{equation}\label{eq:edWarp}
    \begin{split}
    \tilde{\mathbf{v}}_c = ED(\mathbf{v}_c;G) = SE3(\sum\limits_{i\in\mathcal{N}(v_c)}w(\mathbf{x}_i,\mathbf{v}_c)\mathbf{dq}_i)\mathbf{v}_c, 
    \end{split}
    \end{equation}
    where $\mathcal{N}(v_c)$ is a set of node neighbors of $\mathbf{v}_c$, and $w(\mathbf{x}_i,\mathbf{v}_c)=\mathrm{exp}(-\|\mathbf{v}_c-\mathbf{x}_i\|^2_2/(2r_k^2))$ is the influence weight of the $i$-th node $\mathbf{x}_i$ to $\mathbf{v}_c$. The influence radius $r_k$ is set as 0.075m for all the ED nodes. Similarly, $\tilde{\mathbf{n}}_{v_c} = ED(\mathbf{n}_{v_c};G)$ denotes the warped normal of $\mathbf{v}_c$ using the ED motion field $G$.
   
	\subsection{Human Initialization}\label{Sec:initialization}
		
	\begin{figure*}[tbp]
	\centering
 	\includegraphics[width=0.99\textwidth]{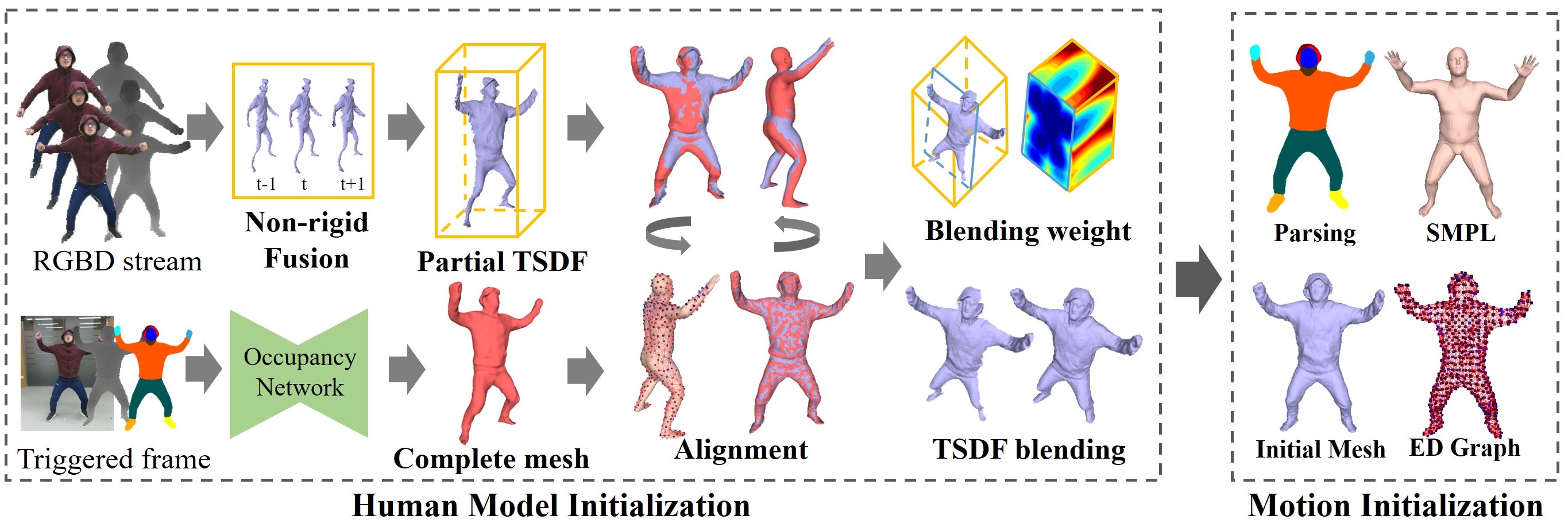}
	\caption{Human model and motion initialization pipeline. Assuming the front-view RGBD input, both a partial TSDF volume and a complete mesh are generated, followed by the alignment and blending operations to obtain a complete human model with fine geometry details, based on which motion is initialized by re-sampling the ED-node-graph and semantic binding.}
	\label{fig_model_init}
    \end{figure*}
	
	Due to complex non-rigid human motions, the good initial human model and motion is critical for us to worry-free focus on human-object interactions. Fortunately,~\cite{RobustFusion2020ECCV} provides us an available robust human performance capture baseline as initialization.
	
	\noindent\textbf{Model initialization.}
	To eliminate the orchestrated self-scanning constraint and the consequent fragile tracking of monocular capture, we propose a model initialization scheme using only the front-view RGBD input.
	As illustrated in Fig.~\ref{fig_model_init}, 
	to generate high-fidelity geometry details, we first utilize the traditional ED-based non-rigid fusion method~\cite{newcombe2015dynamicfusion,FlyFusion} to fuse the depth stream into live partial TSDF volume.
	Once the average accumulated TSDF weight in the front-view voxels reaches a threshold (32 in our setting), a modified RGBD-based PIFu~\cite{PIFU_2019ICCV} network~\cite{RobustFusion2020ECCV} is triggered to generate a watertight mesh. 
	%
	%
	%
	Then, to align the partial TSDF and the complete mesh, we jointly optimize the unique human shape $\bm{\beta}_0$ and pose $\bm{\theta}_0$, as well as the ED motion field $G_0$ from the TSDF volume to the complete mesh as follows:
	\begin{equation}\label{eq:Emot_finalInit}
	\begin{split}
	\boldsymbol{E}_{\mathrm{comp}}(G_0,\bm{\beta}_0,\bm{\theta}_0)=&\lambda_{\mathrm{vd}}\boldsymbol{E}_{\mathrm{vdata}}+\lambda_{\mathrm{md}}\boldsymbol{E}_{\mathrm{mdata}}+\lambda_{\mathrm{bind}}\boldsymbol{E}_{\mathrm{bind}}\\
	&+\lambda_{\mathrm{prior}}\boldsymbol{E}_{\mathrm{prior}}.
	\end{split}
	\end{equation}
	The volumetric data term $\boldsymbol{E}_{\mathrm{vdata}}$ measures the misalignment error between the SMPL and the reconstructed geometry in the partial TSDF volume:
	\begin{equation}\label{eq:Evata}
	\begin{array}{rl}
	\boldsymbol{E}_{\mathrm{vdata}}(\bm{\beta}_0,\bm{\theta}_0)=
	\sum\limits_{\bar{\mathbf{v}}\in{\bar{\textbf{T}}}}\psi(\textbf{D}(W(T(\bar{\mathbf{v}};\bm{\beta}_0, \bm{\theta}_0); \bm{\beta}_0, \bm{\theta}_0)),
	\end{array}
	\end{equation}
	where $\textbf{D}(\cdot)$ takes a point in the canonical volume and returns the bilinear interpolated TSDF, and $\psi(\cdot)$ is the robust Geman-McClure penalty function.
	The mutual data term $E_{\mathrm{mdata}}$ further measures the fitting from both the TSDF volume and the SMPL model to the complete mesh, which is formulated as the sum of point-to-plane distances:
	\begin{equation}\label{eq:Emdata}
	\begin{split}
	\begin{array}{rl}
	\boldsymbol{E}_{\mathrm{mdata}}=&\sum\limits_{(\bar{\mathbf{v}},\mathbf{u})\in{\mathcal{C}}}\psi(\textbf{n}_{{\mathbf{u}}}^{T}(W(T(\bar{\mathbf{v}};\bm{\beta}_0,\bm{\theta}_0))-\mathbf{u})) +\\
	&\sum\limits_{(\tilde{\mathbf{v}}_c,\mathbf{u})\in{\mathcal{P}}}{\psi(\textbf{n}_{{\mathbf{u}}}^{T}(\tilde{\mathbf{v}}_c-\mathbf{u}))},
	\end{array}
	\end{split}
	\end{equation}
	where $\mathcal{C}$ and $\mathcal{P}$ are the correspondence pair sets found via closest searching; $\mathbf{u}$ is a corresponding 3D vertex on the complete mesh and $\textbf{n}_{{\mathbf{u}}}$ is its normal.
	Note that the pose prior term $\boldsymbol{E}_{\mathrm{prior}}$ from \cite{keepitSMPL} penalizes the unnatural poses while the binding term $\boldsymbol{E}_{\mathrm{bind}}$ from \cite{DoubleFusion} constrains both the non-rigid and skeletal motions to be consistent.
	We solve the resulting energy $\boldsymbol{E}_{\mathrm{comp}}$ under the ICP framework, where the non-linear least-squares problem is solved using Levenberg-Marquardt (LM) method with a custom-designed Preconditioned Conjugate Gradient (PCG) solver on GPU~\cite{guo2017real,dou-siggraph2016}.
	Finally, to seamlessly blend both the partial volume and the complete mesh in the TSDF domain,
	we update the voxel as follows:
	\begin{equation}\label{eq:tsdf_udpate}
	\begin{split}
	\textbf{D}(\mathbf{v})\leftarrow{\dfrac{\textbf{D}(\mathbf{v})\textbf{W}(\mathbf{v})+\textbf{d}(\mathbf{v})w(\mathbf{v})}{\textbf{W}(\mathbf{v})+w(\mathbf{v})}}, \\
	\textbf{W}(\mathbf{v})\leftarrow{min(\textbf{W}(\mathbf{v})+w(\mathbf{v}),w_{max})},
	\end{split}
	\end{equation}
	where $\textbf{D}(\mathbf{v})$ and $\textbf{W}(\mathbf{v})$ denote its TSDF value and accumulated weight, respectively, and $w_{max}$ is set as 32 to prevent over-smoothness of geometry during volumetric fusion in Sec.~\ref{Sec: TSDF_fusion} and the corresponding projective SDF value $\textbf{d}(\mathbf{v})$ and the updating weight $\textbf{w}(\mathbf{v})$ are as follows:
	\begin{equation}\label{eq:tsdf_weight}
	\textbf{d}(\mathbf{v})=(\mathbf{u}-\tilde{\mathbf{v}})\textbf{sgn}(\textbf{n}_{{\mathbf{u}}}^{T}(\mathbf{u}-\tilde{\mathbf{v}})), w(\mathbf{v}) = {1}/{(1+\textbf{N}(\mathbf{v}))},
	\end{equation}
	Here, For any 3D voxel $\mathbf{v}$, $\tilde{\mathbf{v}}$ denotes its warped position after applying the ED motion field; $\textbf{N}(\mathbf{v})$ denotes the number of non-empty neighboring voxels of $\mathbf{v}$ in the partial volume which indicates the reliability of the fused geometry, and $\textbf{sgn}(\cdot)$ is the sign function to distinguish positive and negative SDF.
	%
	
	\noindent\textbf{Motion Initialization.}
	The complete model after model initialization provides a reliable initialization for both the human motion and the utilized visual priors.
	As described in Sec.~\ref{Sec:problem}, before the tracking stage, we first re-sample the sparse ED nodes $\{\mathbf{x}_i\}$ on the mesh to form a non-rigid motion field, denoted as $G$, 
	and then we rig the mesh with the pose $\bm{\theta}_0$ from its embedded SMPL model in model initialization and transfer the SMPL skinning weights to the ED nodes $\{\mathbf{x}_i\}$.
	For any 3D point $\mathbf{v}_c$ in the capture volume, let $\tilde{\mathbf{v}}_c$ and $\hat{\mathbf{v}}_c$ denote the warped positions after the embedded deformation and skeletal motion, respectively. 
	Note that the skinning weights of $\mathbf{v}_c$ for the skeletal motion are given by the weighted average of the skinning weights of its knn-nodes.
	%
	To initialize the pose prior, we apply OpenPose~\cite{OpenPose} on the RGBD image to obtain the 2D and lifted 3D joint positions, denoted as $\textbf{P}^{2D}_{l}$ and $\textbf{P}^{3D}_{l}$, respectively, with a detection confidence $\textbf{C}_{l}$.
	Then, we find the closest vertex from the watertight mesh to $\textbf{P}^{3D}_{l}$, denoted as $\mathbf{J}_l$,
	which is the associated marker position for the $l$-th joint.
	To utilize the semantic visual prior, we apply the light-weight human parsing method~\cite{BodyPix} to the triggered RGB image to obtain a human parsing image $L$.
	Then, we project each ED node $\mathbf{x}_i$ into $L$ to obtain its initial semantic label $\mathbf{l}_i$.
	After the motion initialization, inspired by \cite{MonoPerfCap,LiveCap2019tog}, we propose to optimize the motion parameters $G$ and $\bm{\theta}$ in an iterative flip-flop manner to fully utilize the rich motion prior information of the visual cues to capture challenging motions.
	
	\subsection{Scene Decoupling and Object Tracking}\label{Sec: Object}
	Accurate scene decoupling is the premise of robust motion capture that makes full use of the object-specific priors. 
	Otherwise, the wrong segmentation reduces tracking accuracy, and segmentation noise will be fused in the models. However, the semantic segmentation network unavoidably has noise in human-object junction and occlusion, which can not be handled only by the input data. Therefore, we take advantage of our reconstructed human and object models to iteratively refine the segmentation masks to prevent disentanglement uncertainty and maintain temporal consistency.
	As illustrated in Fig.~\ref{fig_overview}, the proposed mask refinement based on initial semantic segmentation provides accurate segmentation of both human and object. Then we can decouple the dynamic scene between rigid object motions and non-rigid human motions to track and reconstruct them. 
	In this subsection, we summarize the mask refinement and the object tracking as follows.
	
	\begin{figure}[tbp]
	\centering
	\subfigure[]{\includegraphics[height=115pt]{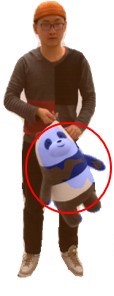}}
	\subfigure[]{\includegraphics[height=115pt]{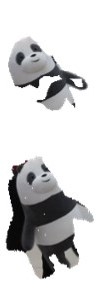}}
	\subfigure[]{\includegraphics[height=115pt]{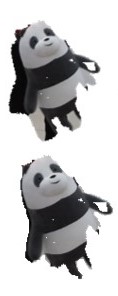}}
	\subfigure[]{\includegraphics[height=115pt]{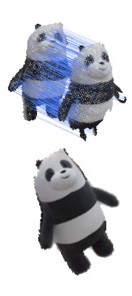}}
	\subfigure[]{\includegraphics[height=115pt]{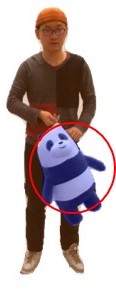}}
	\caption{The results of the proposed mask refinement. (a,e) are the segmentation mask before and after refinement. We sum the original object mask (top of (b)) and projection mask of the previous frame (bottom of (b)) to get the combined mask (top of (c)), then subtract the projection of the human body and remove noise based on point cloud continuity to get a denoised mask (bottom of (c)). Based on the denoised mask, we use Colored-ICP to get the transformation (top of (d)) then use the estimated transformation to get the updated projection mask (bottom of (d)). The detailed iteration procedure is explained in Algorithm. 1.}
	\label{fig_mask_refine}  
	\end{figure}
	
	\noindent\textbf{Mask Refinement.}
	We utilize a semantic segmentation method~\cite{zhao2017pspnet} and the human segmentation from Kinect SDK to get the human and object masks. For operation efficiency, we execute the segmentation network every five frames. At the same time, Kinect SDK provides the human mask for the entire sequence, and the object masks are provided by background separation for the remaining frames. With the human and object masks, we use the masked point cloud for motion tracking and reconstruction. However, wrong segmentation often occurs in the place where people and objects connect. Directly using the coarse segmentation mask leads the unstable tracking and erroneous reconstruction. Therefore, we propose a refinement strategy to obtain the accurate object and human masks, illustrated in Algorithm 1.
    \begin{algorithm}[t]
	\caption{Mask refinement} 		
	\hspace*{0.02in} {\bf Input: } 	
	$M_{o}$,$M_{h}$\\
	\hspace*{0.02in} {\bf Output:} 	
	$M_{o}^{r}$, $M_{h}^{r}$
	\begin{algorithmic}[1]
		\State $M_{o}^{p}=\pi(R_{o})$
		\State $M_{h}^{p}=M_{h}$
		\For{$i=0, i<3$, $i++$} 				
		\State $M_{o}^{r}=M_{o}+M_{o}^{p}$
		\State $M_{o}^{r}=M_{o}^{r}-(dep(M_{h}^{p})<dep(M_{o}^{r}))$
		\State $T_{o}=ICP(M_{o}^{r})$, $M_{o}^{p}=\pi(T_{o}*R_{o})$
		\State $M_{h}^{r}=M_{h}-(dep(M_{o}^{p})<dep(M_{h}))$
		\State $M_{h}^{p}=\pi(track(M_{h}^{r}))$
		\EndFor
		\State \Return $M_{o}^{r}$, $M_{h}^{r}$
	\end{algorithmic}
	\label{alg1}
    \end{algorithm}
	Given the coarse object mask $M_o$ and human mask $M_h$ provided by the segmentation network and Kinect SDK, we aim to get the refined object mask $M_{o}^{r}$ and human mask $M_{h}^{r}$. Here, $R_{o}$ is the reconstructed object model of the previous frame. Function $\pi(\cdot)$ projects the reconstructed model to get the projected mask. $M_{o}^{p}$ and $M_{h}^{p}$ is the projected object and human mask based on the reconstructed models. Due to temporal continuity, the current object mask is similar to the previous frame. The current object mask for tracking is refined as $M_{o}^{r}= M_o+M_{o}^{p}$. Then we remove the human occlusion by comparing the depth (Algorithm 1 Line.5), where function $dep(\cdot)$ returns the depth value for the mask. Based on the refined object mask $M_{o}^{r}$, the transformation between current frame and previous frame $T_{o}$ is solved by optimization $ICP(\cdot)$ and the projected object mask $M_{o}^{p}$ is updated. Then we get the refined human mask $M_{h}^{r}$ by comparing the depth (Algorithm 1 Line.7). Function $track(\cdot)$ returns the update human model based on the refined mask $M_{h}^{r}$. Then the projected human mask is updated by the tracked human model (Algorithm 1 Line.8). Moreover, we also utilize an iteration framework to raise the refinement accuracy. We demonstrate the results of the mask refinement pipeline in Fig.~\ref{fig_mask_refine}. With the mask refinement, we successfully obtain the correct masks.
	
	\noindent\textbf{Object Tracking.}
	To robustly track the objects, we optimize the rigid motions (${\mathbf{T}}=\{T_{i} \},i \in N$) of the corresponding object point clouds under ICP iteration framework as follows:
	\begin{equation}\label{eq:cicp1}
			\vspace{-2pt}
			\boldsymbol{E}_{\mathrm{object}}({\mathbf{T}})=\lambda_{color}\boldsymbol{E}_{\mathrm{color}}+\lambda_{geo}\boldsymbol{E}_{\mathrm{geo}}+\lambda_{sp\_o} \boldsymbol{E}_{\mathrm{sp\_o}}.
	\end{equation}	
	The color term $\boldsymbol{E}_{\mathrm{color}}$ is achieved by the colored point cloud registration~\cite{park2017colored}, which  encourages the color consistency as follows:
	\begin{equation}\label{eq:cicp3}
			\vspace{-2pt}
			\boldsymbol{E}_{\mathrm{color}}=\sum\limits_{i \in N}\sum\limits_{(\mathbf{p},\mathbf{q})\in{\mathcal{R}}}(C_{p}(f(T_{i}\mathbf{q}))-C(\mathbf{q}))^2,
	\end{equation}
	where N is the number of objects, $\mathcal{R}$ is the correspondence pair sets found via closest searching and $\mathbf{p}$, $\mathbf{q}$ are the closest points of frame $t$ and frame $t-1$. Function $C(\cdot)$ returns color of the point $\mathbf{q}$ while $C_{p}(\cdot)$ is a pre-computed function continuously defined on the tangent plane of $\mathbf{p}$ and $f(\cdot)$ is the projection function that projects a 3D point to the tangent plane. The geometry term $\boldsymbol{E}_{\mathrm{geo}}$ encourages the geometry consistency as follows:
	\begin{equation}\label{eq:object_geo}
			\vspace{-2pt}
	\boldsymbol{E}_{\mathrm{geo}}=\sum\limits_{i \in N}\sum\limits_{(\mathbf{p},\mathbf{q})\in{\mathcal{R}}}(\textbf{n}_{{\mathbf{p}}}^{T}(\mathbf{p}-T_{i}\mathbf{q}))^2,
	\end{equation}
	where $\mathbf{n_{p}}$ is the normal of the point $\mathbf{p}$. 
	To generate a healthy spatial relation without implausible interpenetration between human and objects, we introduce an interpenetration term ${E}_{\mathrm{sp\_o}}$ as follows:
	\begin{equation}\label{eq:object_space}
	    \begin{split}
		\vspace{-2pt}
    	\boldsymbol{E}_{\mathrm{sp\_o}}=
    	&\sum\limits_{i \in N}\sum\limits_{\mathbf{p}\in{O_{i}}}\psi(\tau \textbf{D}(T_{i}p))+\\
    	&\sum\limits_{i, j (i\neq j) \in N}\sum\limits_{\mathbf{p}\in{O_{i}}}\psi(\tau \textbf{D}_{oj}(T_{i}p)).
        \end{split}
	\end{equation}
	where $O_{i}$ is the $i$-th object, p is point of $O_{i}$, $\textbf{D}(\cdot)$ is the same as in Eqn.~\ref{eq:Evata}, $\textbf{D}_{oj}(\cdot)$ takes a point in the live TSDF volume of $j$-th object and returns the bilinear interpolated TSDF value, and $\psi(\cdot)$ is the robust Geman-McClure penalty function, and $\tau$ is the indicator function which equals to 1 only if the visited TSDF value is positive (inside the corresponding volume).
	
	\subsection{Robust Human Tracking}\label{Sec:human_track}
	
	\begin{figure*}[tbp]
	\centering
	\includegraphics[width=0.99\textwidth]{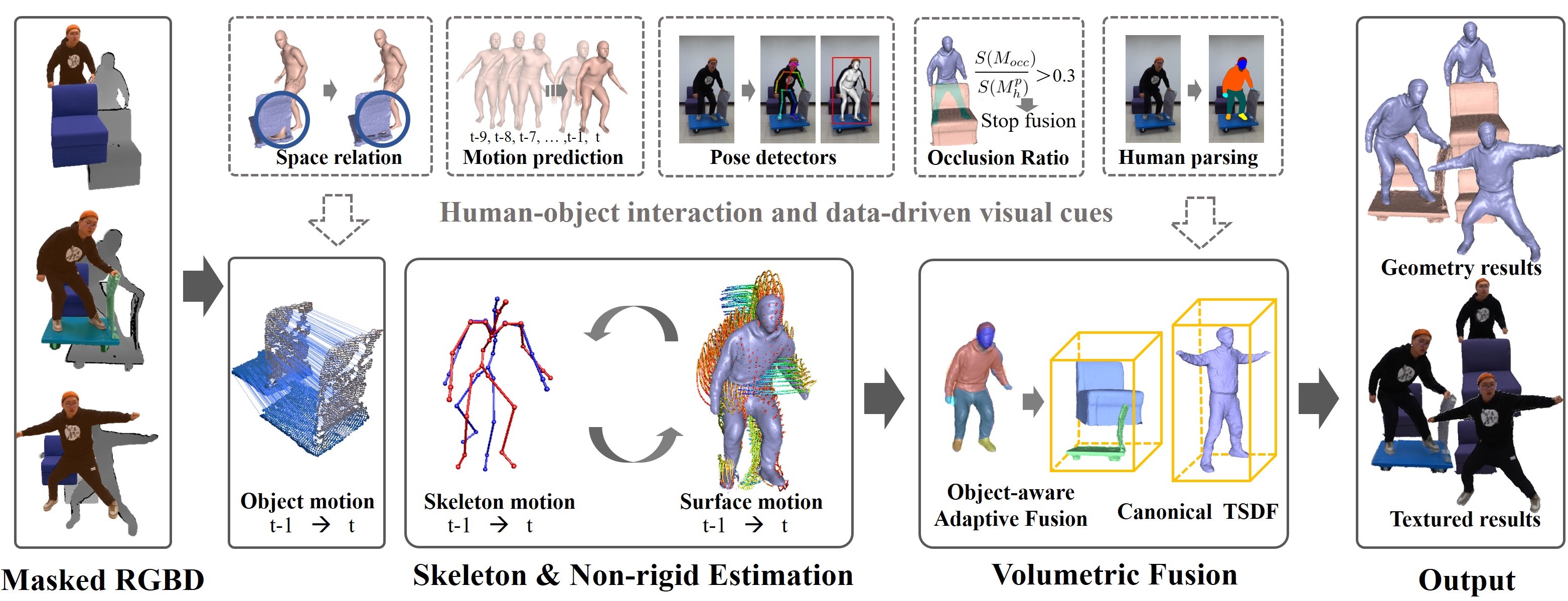}
	\caption{The pipeline of our robust performance capture scheme. Assuming the masked RGBD input, We track the object based on the space relation cue. Then, both skeletal and non-rigid motions are optimized with the associated human-object interaction and data-driven visual cues. Finally, an object-aware adaptive volumetric fusion scheme is adopted to generated 4D models.}
	\label{fig_track_1}
    \end{figure*}
	
    As illustrated in Fig.~\ref{fig_track_1}, we propose a novel performance capture scheme to track challenging human motions under complex human-object interaction scenarios robustly, in which we introduce a spatial relation prior to prevent implausible interactions, data-driven interaction cues to maintain natural motions, especially for those regions under severe human-object occlusions, as well as the human pose, shape and parsing priors to enable re-initialization ability.
	We first optimize the motion fields described in Sec.~\ref{Sec:problem} including both the skeletal pose and surface-sampled ED node-graph in a flip-flop iteration manner.
	
	\noindent\textbf{Skeleton Tracking.}
	During each ICP iteration, we first optimize the skeletal pose $\bm{\theta}$ of the human model, which is formulated as follows:
	\begin{equation}\label{eq:Emot_finalPose}
	\begin{split}
	\boldsymbol{E}_{\mathrm{smot}}(\bm{\theta})=&\lambda_{\mathrm{sd}}\boldsymbol{E}_{\mathrm{sdata}}+\lambda_{\mathrm{pose}}\boldsymbol{E}_{\mathrm{pose}}+\lambda_{\mathrm{prior}}\boldsymbol{E}_{\mathrm{prior}}+
	\\
	&\lambda_{\mathrm{temp}}\boldsymbol{E}_{\mathrm{temp}}+\lambda_{\mathrm{inter}}\boldsymbol{E}_{\mathrm{inter}}.
	\end{split}
	\end{equation}
	
	Here, since human motions have particular patterns in the interactions with objects, we introduce a human-object interaction term $\boldsymbol{E}_{\mathrm{inter}}$  to Eqn.~\ref{eq:Emot_finalPose}, which includes the spacial relation prior, data-driven interaction pose prior, and motion prediction prior to keep natural motions and alleviate the impacts of severe object-human occlusions, formulated as follows:
	\begin{equation}\label{eq:inter}
	\begin{split}
	\boldsymbol{E}_{\mathrm{inter}}=&\lambda_{\mathrm{gmm}}\boldsymbol{E}_{\mathrm{gmm}}+\lambda_{\mathrm{lstm}}\boldsymbol{E}_{\mathrm{lstm}}+\lambda_{\mathrm{sp\_h1}}\boldsymbol{E}_{\mathrm{sp\_h}},
	\end{split}
	\end{equation}
	where $\boldsymbol{E}_{\mathrm{gmm}}$, $\boldsymbol{E}_{\mathrm{lstm}}$ and $\boldsymbol{E}_{\mathrm{sp\_h}}$ are energies of interaction pose prior term, motion prediction prior term and interpenetration term respectively. The interaction pose and motion prior terms come from the experiential data-driven cues, representing the single-frame and temporal priors. At the same time, the interpenetration term represents the spatial prior of human-object interaction.
	The interaction pose prior term resembles the prior term $\boldsymbol{E}_{\mathrm{prior}}$ from \cite{keepitSMPL}. It is based on a Gauss Mixture Model (16 Gaussians) fitted to approximately 200000 human-object interaction temporal poses and formulated as follows:
	\begin{equation}\label{eq:Eprior}
	\begin{array}{rl}
		E_{\mathrm{gmm}} = -log(\sum\limits_{j}w_jN(\bm{\theta};\mu_j,\delta_j)),
	\end{array}
    \end{equation}
	where $w_j$, $\mu_j$ and $\delta_j$ are the mixture weight, the mean, and the variance of $j$-th Gaussian model, respectively.
	Moreover, we train an LSTM predictor to predict the current pose in terms of the poses of the previous nine frames and formulate the motion prediction prior term $\boldsymbol{E}_{\mathrm{lstm}}$ as follows:
	\begin{equation}\label{eq:lstm}
	\begin{split}
	\boldsymbol{E}_{\mathrm{lstm}}=\psi(\bm{\theta}-L(\bm{\theta}_{t-9}, \bm{\theta}_{t-8}, ..., \bm{\theta}_{t-1})),
	\end{split}
	\end{equation}
	where $\psi(\cdot)$ is the robust Geman-McClure penalty function; $\bm{\theta}_{i}, (i = t-9, t-8, ...,t-1)$ are the skeleton poses of the previous 9 frames; $L(\cdot)$ is the LSTM prediction fuction. 
	The interpenetration term $\boldsymbol{E}_{\mathrm{sp\_h}}$ prevents unphysical intersections from human to objects in space dimension:
	\begin{equation}\label{eq:human_space}
	\begin{split}
	\boldsymbol{E}_{\mathrm{sp\_h}}=\sum\limits_{\mathbf{v}\in{\textbf{T}}}\psi(\tau \textbf{D}_o(W(T(\mathbf{v}; \bm{\beta}, \bm{\theta}); \bm{\theta})),
	\end{split}
	\end{equation}
    where $\textbf{D}_o(\cdot)$ takes a point in the live TSDF volume of the object and returns the bilinear interpolated TSDF value, and $\psi(\cdot)$ is the robust Geman-McClure penalty function, $\tau$ is the indicator function which equals to 1 only if the TSDF value for the object of the vertex $\mathbf{v}$ on SMPL is positive (inside the object volume).

	Besides, we also introduce the pose term $\boldsymbol{E}_{\mathrm{pose}}$ in~\cite{RobustFusion2020ECCV} and update its pose detectors to better encourage the skeleton to match the detections obtained by CNN from the RGB image, including the 2D position $\textbf{P}^{2D}_{l}$, lifted 3D position $\textbf{P}^{3D}_{l}$ and the pose parameters $\bm{\theta}_d$ from  OpenPose~\cite{OpenPose} and EFT~\cite{joo2020eft} (HMR~\cite{HMR18} in \cite{RobustFusion2020ECCV}):
	\begin{equation}\label{eq:Esmot_pose}
	\begin{split}
	\boldsymbol{E}_{\mathrm{pose}}=&\psi(\Phi^{T}(\bm{\theta}-\bm{\theta}_d)) + \sum_{l=1}^{N_J}\phi(l)(\|\pi(\hat{\mathbf{J}}_l)-\textbf{P}^{2D}_{l}\|_2^2 +
	\\
	&\|\hat{\mathbf{J}}_l-\textbf{P}^{3D}_{l}\|_2^2),
	\vspace{-5pt}
	\end{split}
	\end{equation}
	where $\psi(\cdot)$ is the robust Geman-McClure penalty function; $\hat{\mathbf{J}}_l$ is the warped associated 3D position and $\pi(\cdot)$ is the projection operator. 
	The indicator $\phi(l)$ equals to 1 if the confidence $\textbf{C}_{l}$ for the $l$-th joint is larger than 0.5, while $\Phi$ is the vectorized representation of $\{\phi(l)\}$.

	Finally, among the other terms, $\boldsymbol{E}_{\mathrm{sdata}}$ measures the point-to-plane misalignment error between the warped geometry in the TSDF volume and the depth input: 
	\begin{equation}\label{eq:Esmot_data}
	\boldsymbol{E}_{\mathrm{sdata}}=\sum\limits_{(\mathbf{v}_c,\mathbf{u})\in{\mathcal{P}}}{\psi(\textbf{n}_{{\mathbf{u}}}^{T}(\hat{\mathbf{v}}_c-\mathbf{u}))},
	\end{equation}
	where $\mathcal{P}$ is the corresponding set found via a projective searching; $\mathbf{u}$ is a sampled point on the depth map while $\mathbf{v}_c$ is the closet vertex on the fused surface; 
	the temporal term $\boldsymbol{E}_{\mathrm{temp}}$ encourages coherent deformations by constraining the skeletal motion to be consistent with the previous ED motion:
	\begin{equation}\label{eq:Esmot_temp}
	\boldsymbol{E}_{\mathrm{temp}}=\sum_{\mathbf{x}_i}\|\hat{\mathbf{x}}_i - \tilde{\mathbf{x}}_i\|_2^2,
	\vspace{-4pt}
	\end{equation}
	where $\tilde{\mathbf{x}}_i$ is the warped ED node using non-rigid motion from previous iteration;
	and the prior term $\boldsymbol{E}_{\mathrm{prior}}$ from \cite{keepitSMPL} penalizes the unnatural poses.
	
	\noindent\textbf{Surface Tracking.}
	To capture realistic non-rigid deformation defined by ED-node graph $G$, on top of the skeleton tracking result, we solve the surface tracking energy as follows:
	\begin{equation}\label{eq:Emot_finalED}
	\vspace{-2pt}
	\begin{split}
	\boldsymbol{E}_{\mathrm{emot}}(G)=&\lambda_{\mathrm{ed}}\boldsymbol{E}_{\mathrm{edata}}+\lambda_{\mathrm{sp\_h2}}\boldsymbol{E}_{\mathrm{sp\_h}}+\lambda_{\mathrm{reg}}\boldsymbol{E}_{\mathrm{reg}}+\\
	&\lambda_{\mathrm{temp}}\boldsymbol{E}_{\mathrm{temp}}.
	\end{split}
	\end{equation}

	Here the dense data term $E_{\mathrm{edata}}$ jointly measures the dense point-to-plane misalignment and the sparse landmark-based projected error:
	\begin{equation}\label{eq:Eemot_data}
	\vspace{-2pt}
	\boldsymbol{E}_{\mathrm{edata}}=\sum\limits_{(\mathbf{v}_c,\mathbf{u})\in{\mathcal{P}}}{\psi(\textbf{n}_{{\mathbf{u}}}^{T}(\tilde{\mathbf{v}}_c-\mathbf{u}))} + \sum_{l=1}^{N_J}\phi(l)\|\pi(\tilde{\mathbf{J}}_l)-\textbf{P}^{2D}_{l}\|_2^2,
	\end{equation}
	where $\tilde{\mathbf{J}}_l$ is the warped associated 3D joint of the $l$-th joint in the fused surface.
	The interpenetration term $\boldsymbol{E}_{\mathrm{sp\_h2}}$ is as follows:
	\begin{equation}\label{eq:human_space_2}
	\begin{split}
		\boldsymbol{E}_{\mathrm{sp\_h2}}=\sum\limits_{\mathbf{v}\in{\textbf{T}}}\psi(\tau \textbf{D}_o(\tilde{\mathbf{v}})),
	\end{split}
   \end{equation}
	note that the interpenetration term is associated with a smaller weight : $\mathrm{sp\_h2}=\mathrm{sp\_h1}/10$.
	The regularity term $\boldsymbol{E}_{\mathrm{reg}}$ from \cite{DoubleFusion} produces locally as-rigid-as-possible (ARAP) motions to prevent over-fitting to depth inputs.
	Besides, the $\hat{\mathbf{x}}_i$ after the skeletal motion in the temporal term $\boldsymbol{E}_{\mathrm{temp}}$ as formulated above is fixed during current optimization.
	
	Both the pose and non-rigid optimizations in Eqn.~\ref{eq:Emot_finalPose} and Eqn.~\ref{eq:Emot_finalED} are solved using LM method with the same PCG solver on GPU~\cite{guo2017real,dou-siggraph2016}.
	%
	%
	Once the confidence $\textbf{C}_{l}$ reaches 0.9 and the projective error $\|\pi(\tilde{\mathbf{J}}_l)-\textbf{P}^{2D}_{l}\|_2^2$ is larger than 5.0 for the $l$-th joint, the associated 3D position $\mathbf{J}_l$ on the fused surface is updated via the same closest searching strategy of the initialization stage. When there is no human detected in the image, our whole pipeline will be suspended until the number of detected joints reaches a threshold (10 in our setting).

	\subsection{Object-aware Reconstruction}\label{Sec: TSDF_fusion}
	
    After the above optimization, we separately fuse the masked depth into the respective canonical TSDF volume of the human and objects with occlusion analysis and human semantic cue to temporally update the geometric details. Note that each voxel in canonical space is updated using Eqn.~\ref{eq:tsdf_udpate}, while  updating weight $\textbf{w}(\mathbf{v})$ is different between human and objects.

	For human reconstruction, we first discard the voxels which are collided or warped into invalid input. Then, to avoid deteriorated fusion caused by challenging motion, an effective adaptive fusion strategy as shown in Fig.~\ref{fig_track_1} is proposed to model semantic motion tracking behavior.
	%
	To this end, we apply the human parsing method~\cite{BodyPix} to the current RGB image to obtain a human parsing image $L$.
	For each ED node $\mathbf{x}_i$, recall that $\mathbf{l}_i$ is its associated semantic label during initialization while $L(\pi(\tilde{\mathbf{x}}_i)$ is current corresponding projected label.
	Then, for any voxel $\mathbf{v}$, we formulate its updating weight $\textbf{w}(\mathbf{v})$ as follows:
	\begin{equation}\label{eq:adapt_fusion}
		\mathbf{w}(\mathbf{v}) =\exp(\frac{-\|\Phi^{T}(\bm{\theta}^*-\bm{\theta}_d)\|_2^2}{2\pi})\sum\limits_{i\in\mathcal{N}(v_c)}\frac{\varphi(\mathbf{l}_i,L(\pi(\tilde{\mathbf{x}}_i)))}{|\mathcal{N}(v_c)|},
	\end{equation}
	where $\bm{\theta}^*$ is the optimized pose; $\mathcal{N}(v_c)$ is the collection of the knn-nodes of $\mathbf{v}$; $\varphi(\cdot,\cdot)$ denote an indicator which equals to 1 only if the two input labels are the same.
	Note that such a robust weighting strategy measures the tracking performance based on the human pose and semantic priors.
	Then, $\textbf{w}(\mathbf{v})$ is set to be zero if it is less than a truncated threshold (0.2 in our setting), to control the minimal integration and further avoid deteriorated fusion of severe tracking failures. 
	$M_{occ}=(dep(M_{h}^{p})>dep(M_{o}^{p}))$ is the mask of human occluded by object.	Function $S(\cdot)$ returns the pixel number of mask. When the severe object-human occlusion occurs as $S(M_{occ})/S(M_{h}^{p})$ is bigger than a threshold (0.3 in our setting), we also set $\textbf{w}(\mathbf{v})$ to zero in case deteriorated fusion caused by false segmentation results and tracking errors caused by occlusions. 
	%
	
	As for object reconstruction, $R_{in}$ is the root mean square error (RMSE) of all inlier correspondences in the ICP framework. TSDF fusion is performed every five frames based on Eqn.~\ref{eq:tsdf_udpate} only if $R_{in}$ is less than a certain value (0.003 in our setting), in which the updating weight $\textbf{w}(\mathbf{v})$ is formulated as $\textbf{w}(\mathbf{v})=\frac{0.0048}{R_{in}+0.0024}$. Again, similar to human volumetric fusion, once the human occludes an object, we stop the TSDF fusion for the whole object. Finally, the dynamic atlas scheme~\cite{UnstructureLan} and per-vertex color fusion are adopted to obtain 4D textured reconstruction results for human and objects, respectively.
	
	\begin{figure*}[htbp]
	\centering
	\includegraphics[width=1.0\textwidth]{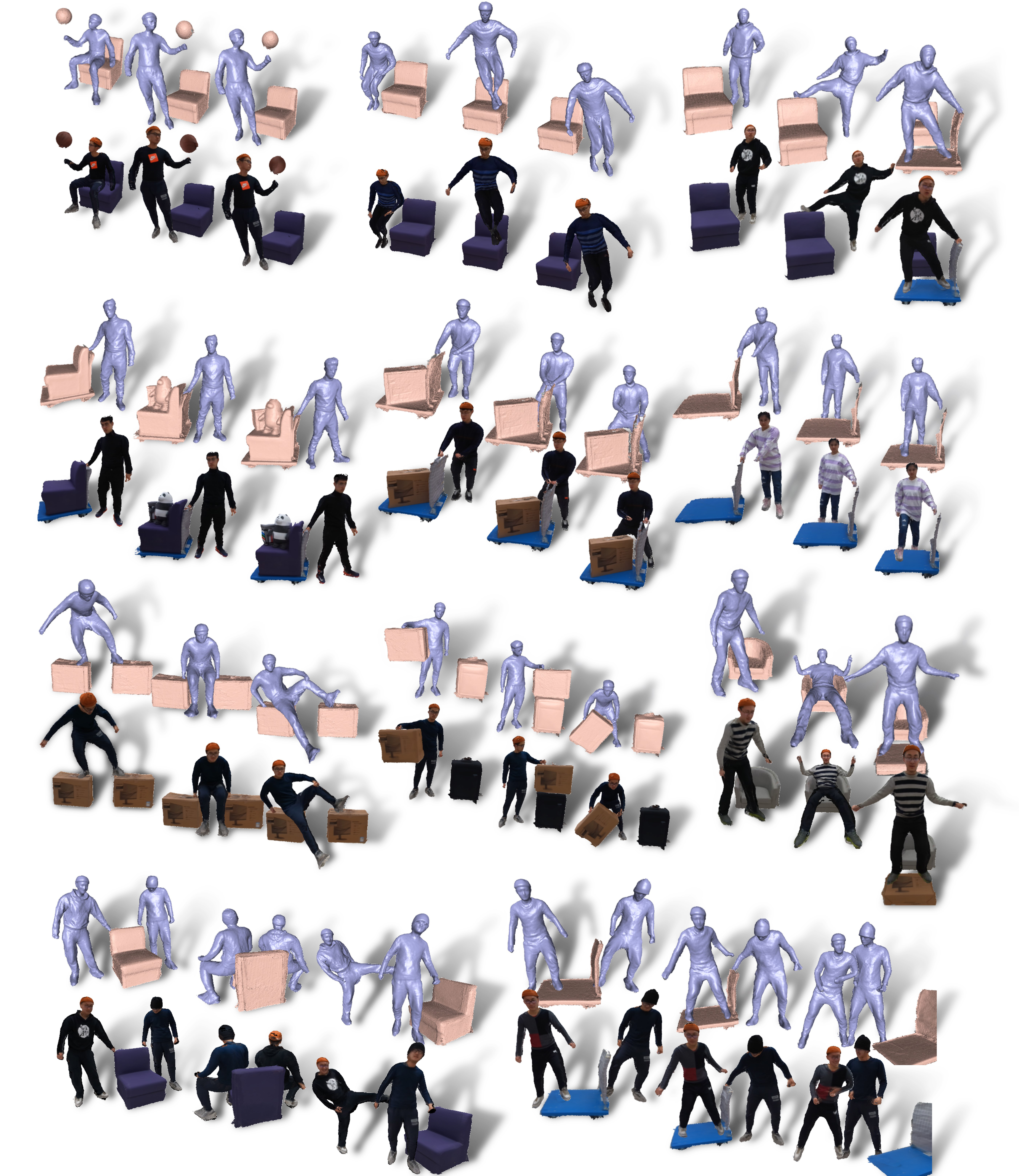}
	\caption{4D human and object reconstructed results of the proposed RobustFusion system, and the interacted objects include a sofa, a cart, two cartons, a piece of luggage, a chair, and a toy.}
	\label{fig_all_result}
    \end{figure*}
	
	\section{Experimental Results}\label{sec:Experiement}
	\label{Experiement}
	In this section, we first report the performance and the main parameters of our RobustFusion system.
Then we compare with the state-of-the-art methods and evaluate the technical contributions of our RobustFusion both qualitatively and quantitatively on a variety of challenging scenarios in Sec.~\ref{Sec:compare} and Sec.~\ref{Sec:evaluation}, respectively.
Fig.~\ref{fig_all_result} demonstrates the results of RobustFusion, where both the challenging motions with human-object interactions and the fine geometry and texture details are faithfully captured. 
Our approach can even faithfully reconstruct the interaction scenarios with multiple performers and various objects (see the last row of Fig.~\ref{fig_all_result}).
Please also kindly refer to the supplemental video for the sequential 4D reconstruction results.

\subsection{Performance}
We run our experiments on a PC with an NVIDIA GeForce GTX TITAN Xp GPU and an Intel Core i7-7700K CPU. 
Our human initialization takes 15 s, and the following robust performance capture pipeline runs at an average of 135 ms per frame, where the visual priors collecting takes 97 ms, the robust human-object tracking takes around 21 ms with 4 ICP iteration and 17 ms on average for all the remaining computations.
Note that the semantic segmentation network and volumetric fusion for objects are executed every five frames.
In all experiments, we use the following empirically determined parameters: $\lambda_{vd} = 1.0$, $\lambda_{md} = 2.0$, $\lambda_{bind} = 1.0$, $\lambda_{prior} = 0.01$, $\lambda_{color} = 0.1$, $\lambda_{geo} = 0.9$, $\lambda_{sp\_o} = 1.0$, $\lambda_{sd} = 4.0$, $\lambda_{pose} = 2.0$, $\lambda_{temp} = 1.0$, $\lambda_{inter} = 1.0$, $\lambda_{gmm} = 0.02$, $\lambda_{lstm} = 0.1$, $\lambda_{sp\_h1} = 2.0$, $\lambda_{sp\_h2} = 0.2$, $\lambda_{ed} = 4.0$ and $\lambda_{reg} = 5.0$.
For the ED model, we use the four nearest node neighbors for ED warping and the eight nearest node neighbors to construct the ED graph. For the TSDF voxel, the size is set as 4 mm in each dimension.

\begin{figure*}[t]
	\centering
	\subfigure[]{\includegraphics[height=210pt]{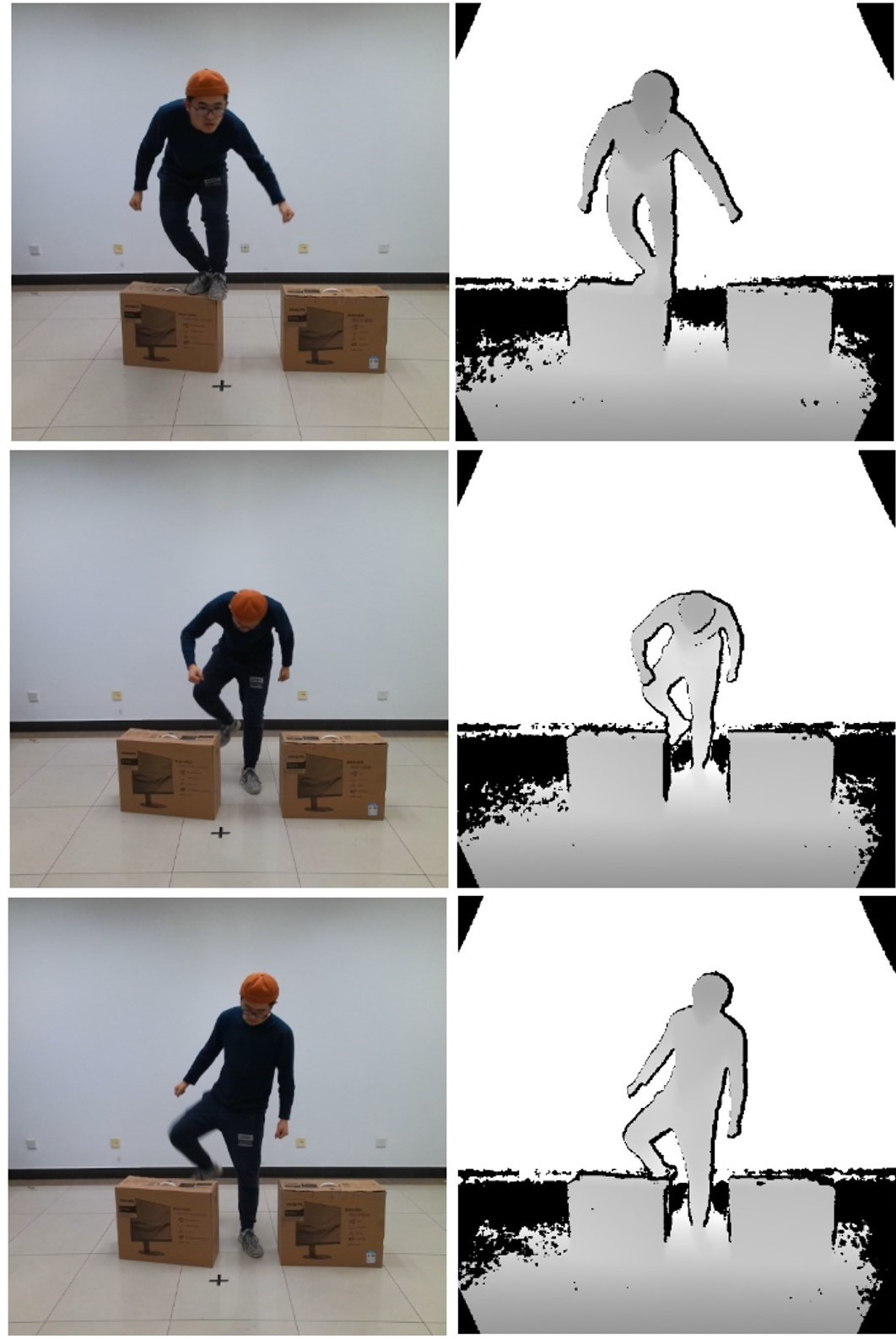}}
	\subfigure[]{\includegraphics[height=210pt]{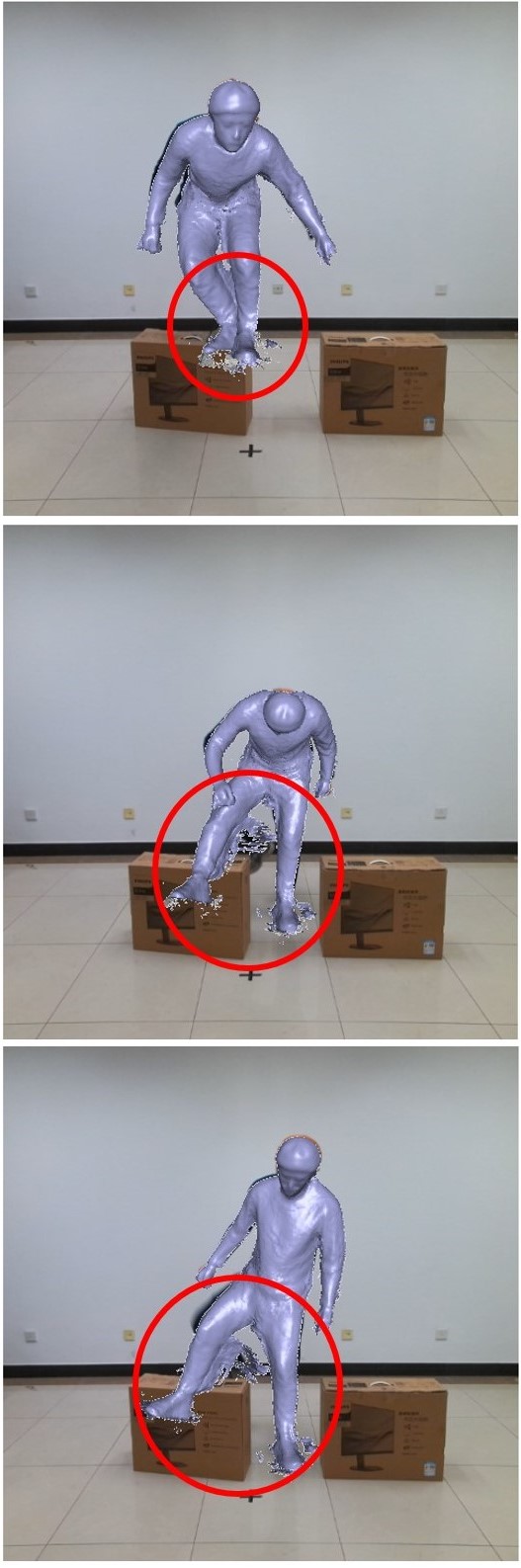}}	             \subfigure[]{\includegraphics[height=210pt]{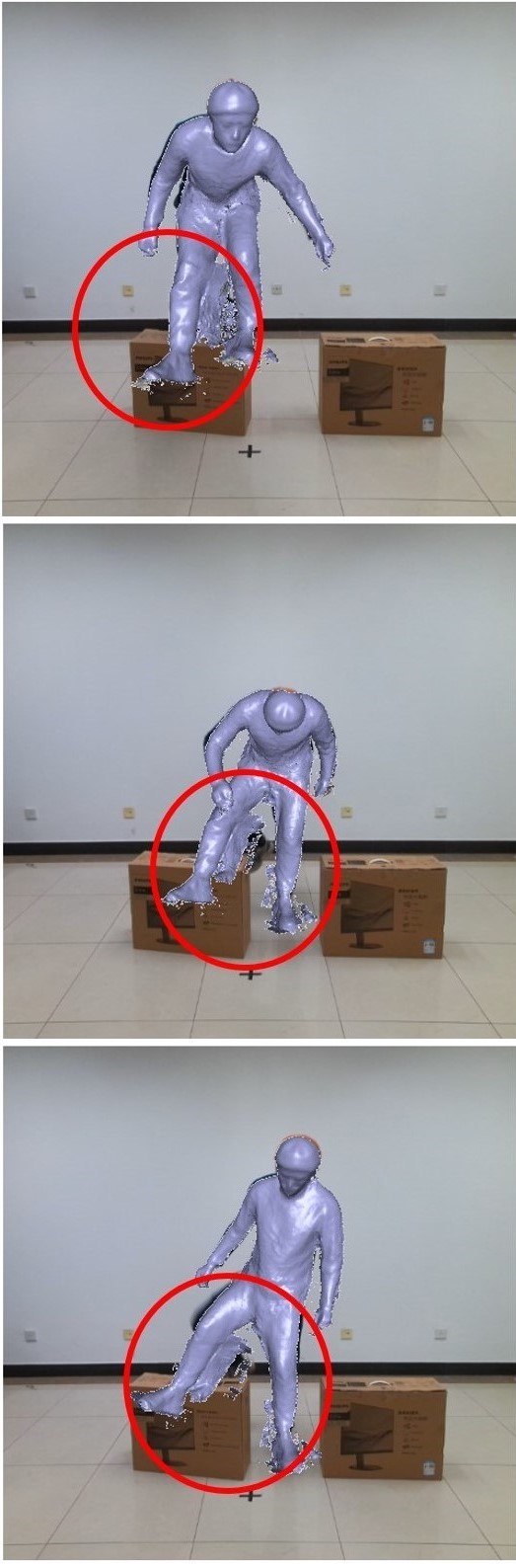}}
	\subfigure[]{\includegraphics[height=210pt]{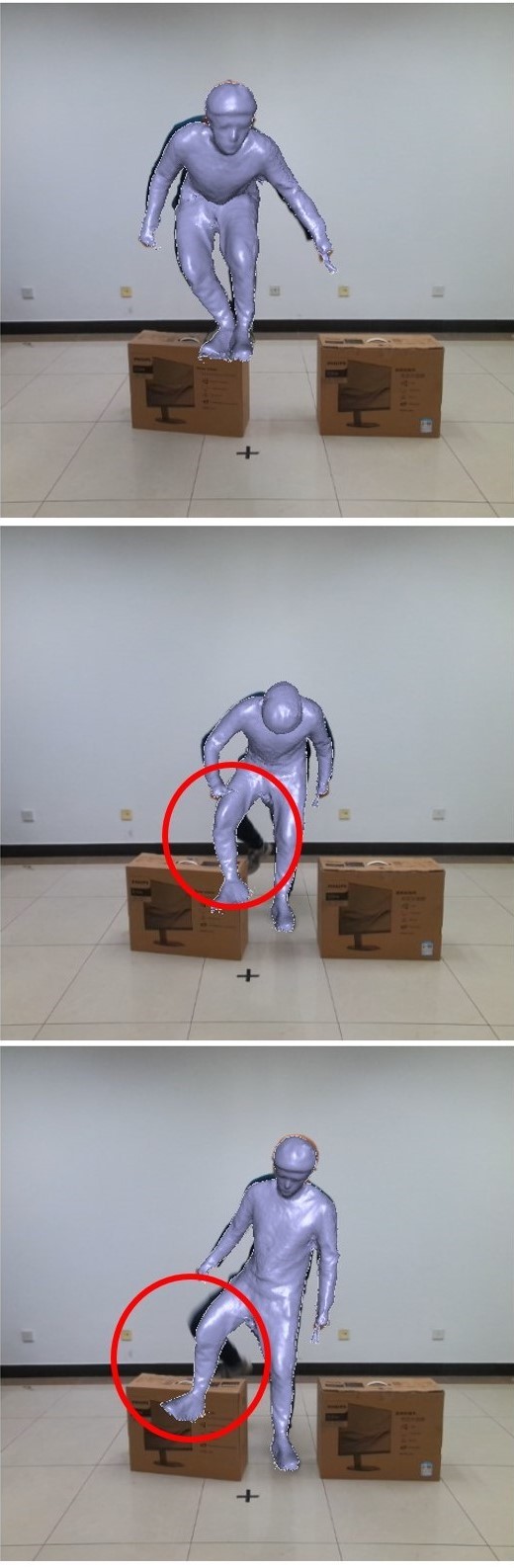}}
	\subfigure[]{\includegraphics[height=210pt]{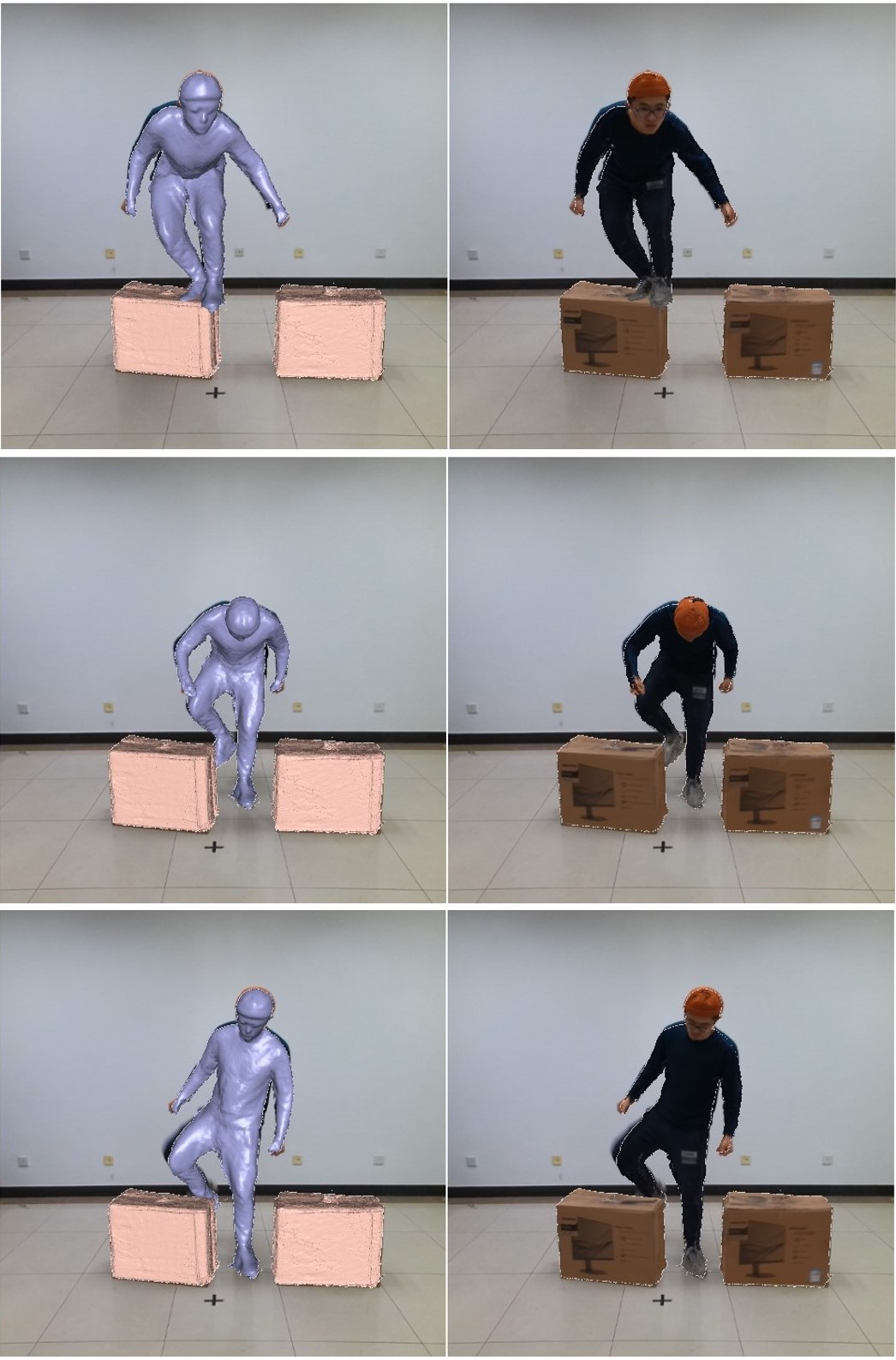}}
	\caption{Qualitative comparison.(a) is the reference RGBD images. (b-d) are the geometry results of DoubleFusion~\cite{DoubleFusion},  UnstructuredFusion~\cite{UnstructureLan} and RobustFusion(Conf.)~\cite{RobustFusion2020ECCV}, respectively. (e) is the geometry/texture results of our Proposed RobustFusion.}
	\label{fig_compare_1}
\end{figure*}

\begin{table}[tbp]
	\footnotesize
	\newcommand{\tabincell}[2]{\begin{tabular}{@{}#1@{}}#2\end{tabular}}
	\renewcommand{\arraystretch}{1.5}
	\addtolength{\tabcolsep}{-2.5pt}
	\centering
	\caption{Average projective numerical errors (mm) of our captured sequences for the concerned methods: DoubleFusion \cite{DoubleFusion}, UnstructuredFusion \cite{UnstructureLan}, RobustFusion(Conf.)~\cite{RobustFusion2020ECCV} and our methods, where the corresponding sequences can refer to the supplementary video.}

    \begin{tabular}{p{3.4 cm}p{1.0cm}p{1.0cm}p{1.0cm}p{1.0cm}}
		\hline
		Human-object Interactions & \tabincell{c}{\cite{DoubleFusion}} & \tabincell{c}{\cite{UnstructureLan}}  & \tabincell{c}{\cite{RobustFusion2020ECCV}} & \tabincell{c}{Ours}\\
		\hline
		{\small{\textit{with luggage \& chair}}}& 29.66    & 28.34  & 22.32  & 17.74 \\ 
		\hline
		{\small{\textit{with two cartons}}}& 16.56   & 11.38  & 8.37  & 7.61 \\
		\hline
		{\small{\textit{dragging things}}}& 9.11   & 8.56  & 6.45  & 4.96 \\   
		\hline
		{\small{\textit{rotating a chair}}}& 17.45   & 15.10  & 10.34  & 8.61 \\ 
		\hline
		{\small{\textit{with a luggage}}}& 18.14   & 13.24  & 9.42  & 7.26 \\		
		\hline
		{\small{\textit{with luggage \& carton (1)}}}& 14.35   & 10.57  & 8.51  & 7.87 \\		
		\hline
		{\small{\textit{with luggage \& carton (2)}}}& 18.48   & 13.82  & 10.72  & 8.50 \\
		\hline
		{\small{\textit{with a cart (girl)}}}& 17.75   & 12.47  & 9.49  & 8.87 \\
		\hline
		{\small{\textit{with cart \& carton (1)}}}& 17.83   & 13.27  & 10.34  & 8.80 \\
		\hline
		{\small{\textit{with cart \& carton (2)}}}& 12.93   & 7.91  & 5.48  & 5.05 \\
		\hline
		{\small{\textit{with a sofa}}}& 11.64   & 7.66  & 7.11  & 6.60 \\
		\hline
		{\small{\textit{with a cart (boy)}}}& 23.14   & 18.96  & 15.55  & 15.04 \\
		\hline
		{\small{\textit{with backpack \& toy}}}& 34.34   & 32.12  & 28.34  & 23.37 \\
		\hline
	\end{tabular}
	\label{table:allcomp}
\end{table}

\subsection{Comparison}\label{Sec:compare}
For throughout comparison, we compare our RobustFusion against the state-of-the-art methods in this subsection, including DoubleFusion~\cite{DoubleFusion},  UnstructuredFusion~\cite{UnstructureLan}, HybridFusion~\cite{HybridFusion}, RobustFusion(Conf.)~\cite{RobustFusion2020ECCV} and POSEFusion~\cite{li2021posefusion} both qualitatively and quantitatively.

\noindent\textbf{Qualitative Comparison.}
These state-of-the-art methods are restricted to human reconstruction without modeling human-object interactions, and UnstructuredFusion~\cite{UnstructureLan} is a multi-view method. 
For a fair comparison of dynamic reconstruction at the scenes with objects, we test the above state-of-the-art methods on the same refined segmentation results of the human in our setting and modify UnstructuredFusion~\cite{UnstructureLan} into the monocular setting by removing their online calibration stage.

\begin{figure}[htb]
	\centering
	\includegraphics[width=0.95\linewidth]{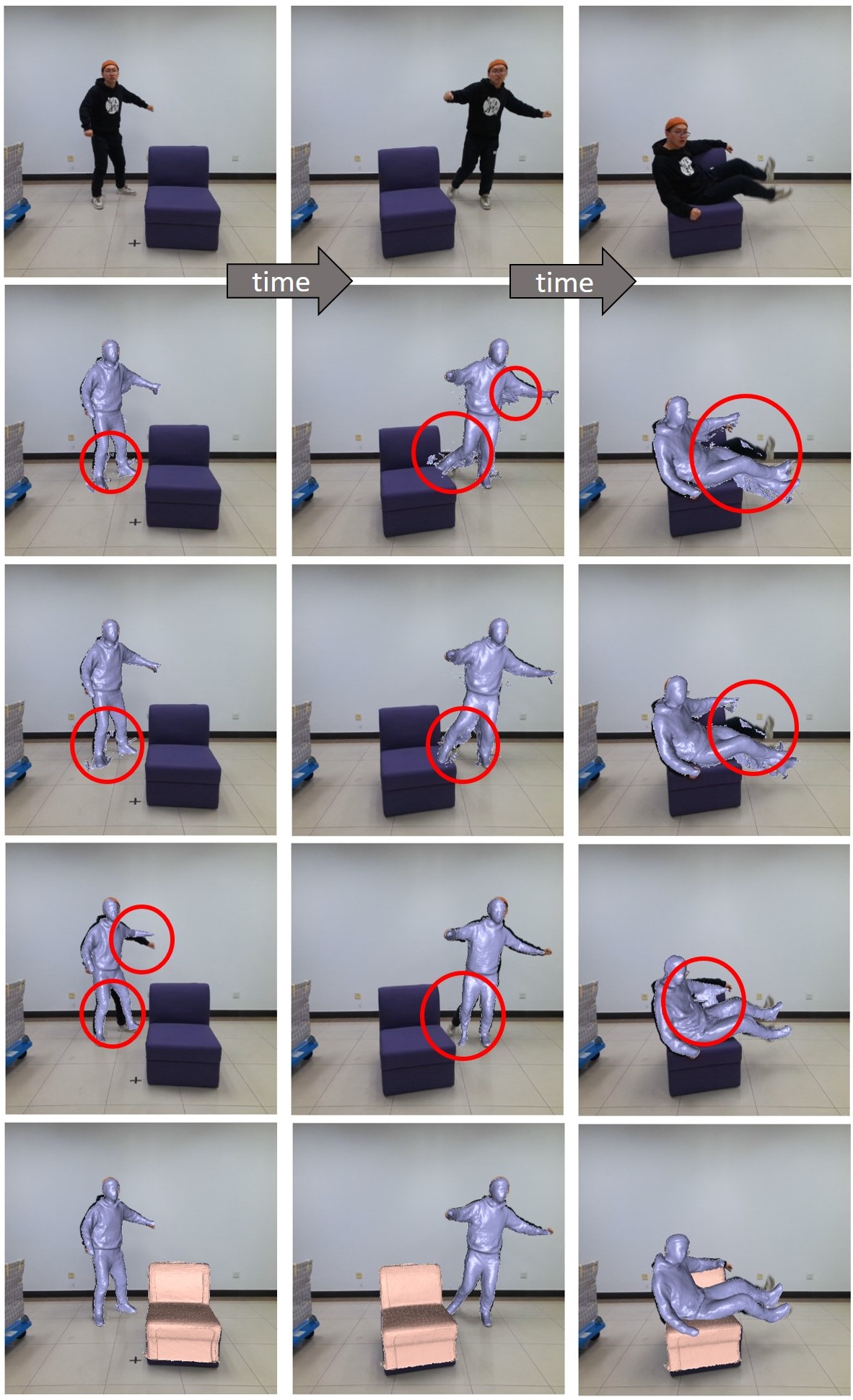}
	\caption{Qualitative comparison. The first row is the reference RGB images. The second to fifth row are the geometry results of DoubleFusion~\cite{DoubleFusion},  UnstructuredFusion~\cite{UnstructureLan} and POSEFusion~\cite{li2021posefusion} and our method,  respectively.}
	\label{fig_com_posefusion}
\end{figure}

The qualitative comparison of our approach against DoubleFusion~\cite{DoubleFusion}, single-view UnstructuredFusion~\cite{UnstructureLan} and RobustFusion(Conf.)~\cite{RobustFusion2020ECCV} is as shown in Fig.~\ref{fig_compare_1}.
Both DoubleFusion~\cite{DoubleFusion} and UnstructuredFusion~\cite{UnstructureLan} suffer from the fast human motions and the severe occlusions due to the human-object interactions. 
Moreover, without a complete model due to the lack of orchestrated self-circling motions, they tend to integrate erroneous surfaces at the newly fused region.
With the aid of various visual priors, RobustFusion(Conf.)~\cite{RobustFusion2020ECCV} is more robust to the fast motions but still suffers from severe occlusions, leading to wrong tracking results in the limb regions. 
In contrast, benefit from our robust human tracking scheme based on data-driven interaction and visual cues, our approach achieves significantly more robust tracking results, especially for challenging occluded and fast motions.
Besides, we compare against the latest volumetric method POSEFusion~\cite{li2021posefusion} in Fig.~\ref{fig_com_posefusion}, which combines implicit inference network with a key-frame selection strategy to capture details in invisible regions.
As shown in Fig.~\ref{fig_com_posefusion}, our approach can achieve more accurate human tracking and visually pleasant reconstruction results with the aid of human-object interaction cues.
Nevertheless, our approach can faithfully reconstruct both the humans and objects in the interaction scenarios, which is unseen in the previous monocular fusion approaches.

\begin{figure}[htb]
	\centering
	\subfigure[]{\includegraphics[height=135pt]{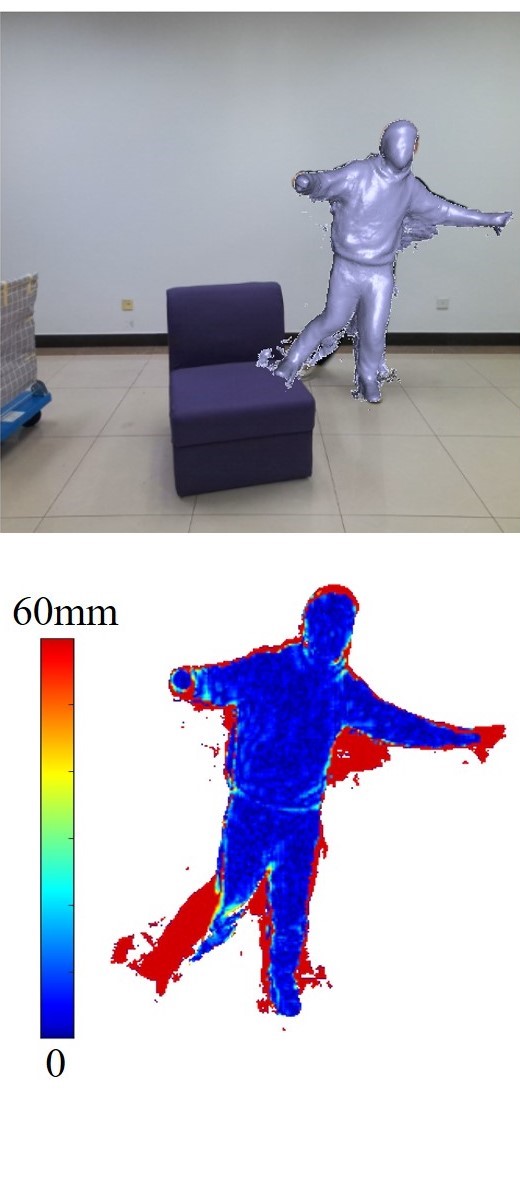}}
	\subfigure[]{\includegraphics[height=135pt]{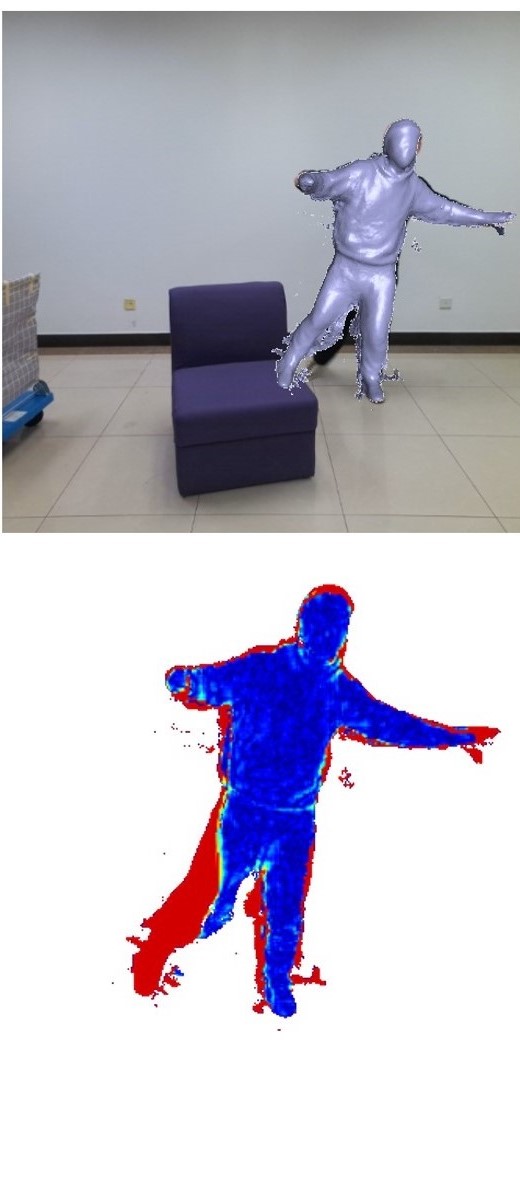}}	 \subfigure[]{\includegraphics[height=135pt]{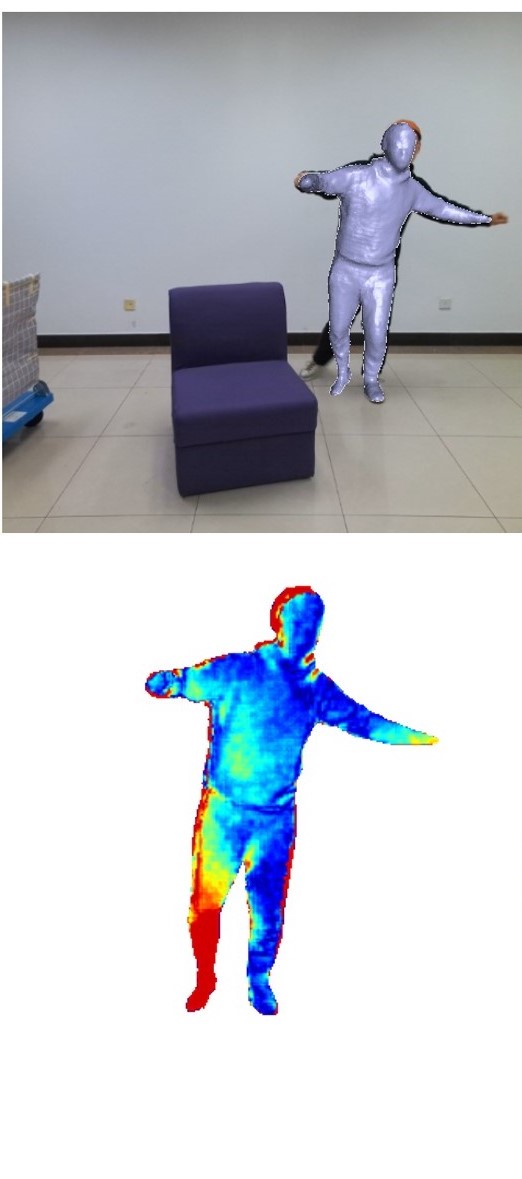}}
	\subfigure[]{\includegraphics[height=135pt]{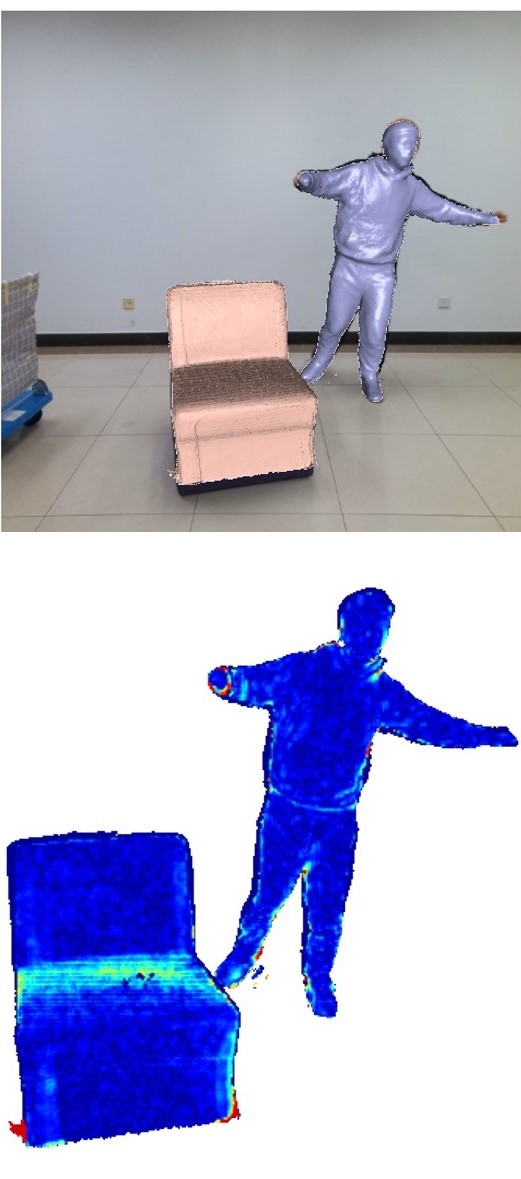}}
	\caption{Qualitative comparison. (a-d) are the geometry results of DoubleFusion~\cite{DoubleFusion},  UnstructuredFusion~\cite{UnstructureLan}, POSEFusion~\cite{li2021posefusion} and our Proposed RobustFusion, respectively. The color-coded maps in bottom row indicate the projective errors.}
	\label{fig_compare_2}
\end{figure}

\noindent\textbf{Quantitative Comparison.}
For quantitative comparison, we first utilize the average projective numerical metric. Specifically, we render the reconstructed result to a depth map in the camera view and compute its MAE (Mean Absolute Error) by taking the depth input as the reference only in the intersection between the rendered surface and the human depth. 
Note that even without ground truth reconstruction, this MAE metric encodes the reconstruction error for the non-rigid motion capture process of each method, providing a reliable quantitative comparison.
We only compute MAE in the human regions for a fair comparison since previous methods cannot reconstruct objects.
Tab.~\ref{table:allcomp} demonstrates the MAE of different sequences in our experiments, in which our method leads to considerably less error, i.e., 10.02 mm average MAE, compared with 18.57 mm of DoubleFusion~\cite{DoubleFusion}, 14.88 mm of UnstructuredFusion~\cite{UnstructureLan} and 11.73 mm of RobustFusion(Conf.)~\cite{RobustFusion2020ECCV}. 
Moreover, Fig.~\ref{fig_compare_2} demonstrates that our method achieves high-quality reconstruction results with less accumulated artifacts, using the corresponding sequence ``Human-object interactions with a sofa'' in Tab.~\ref{table:allcomp}.
Note that our MAE for this sequence is 6.60 mm, compared favorably with 17.24 mm for the reconstructed results provided by  POSEFusion~\cite{li2021posefusion}.
These quantitative comparisons above reveal the effectiveness of our method for more robust and accurate human motion tracking and reconstruction.

\begin{figure}[tbp]
	\centering
	\subfigure[]{\includegraphics[height=110pt]{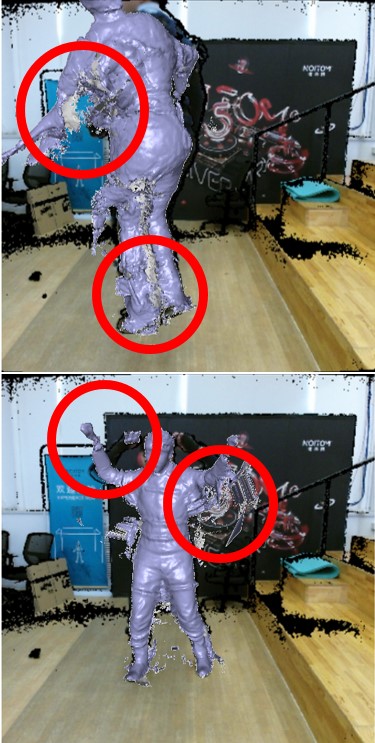}}
	\subfigure[]{\includegraphics[height=110pt]{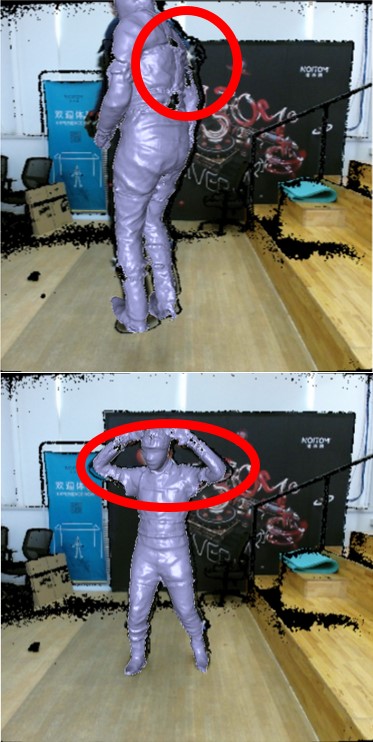}}	\subfigure[]{\includegraphics[height=110pt]{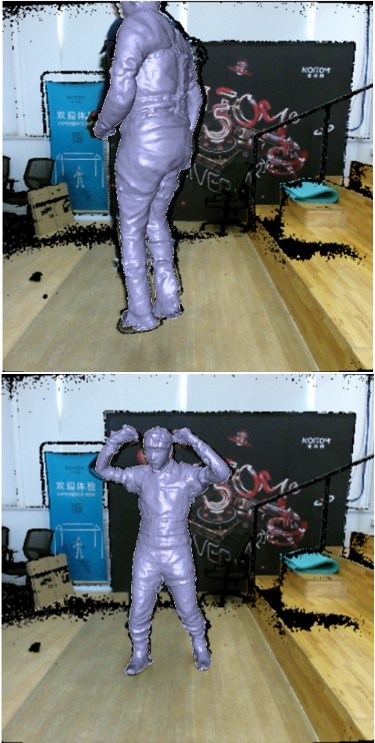}}
	\subfigure[]{\includegraphics[height=110pt]{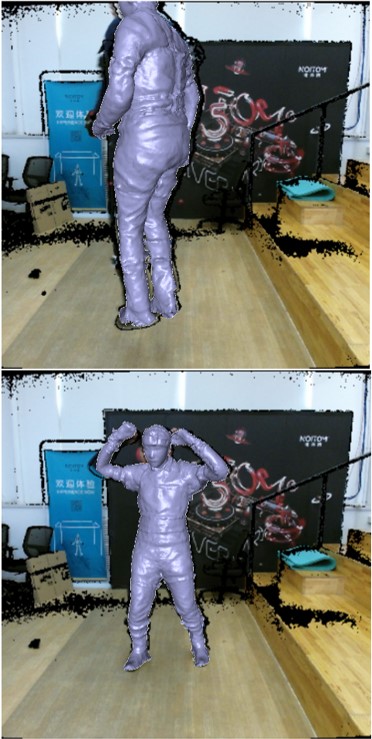}}
	
	\subfigure[]{\includegraphics[width=0.93\linewidth]{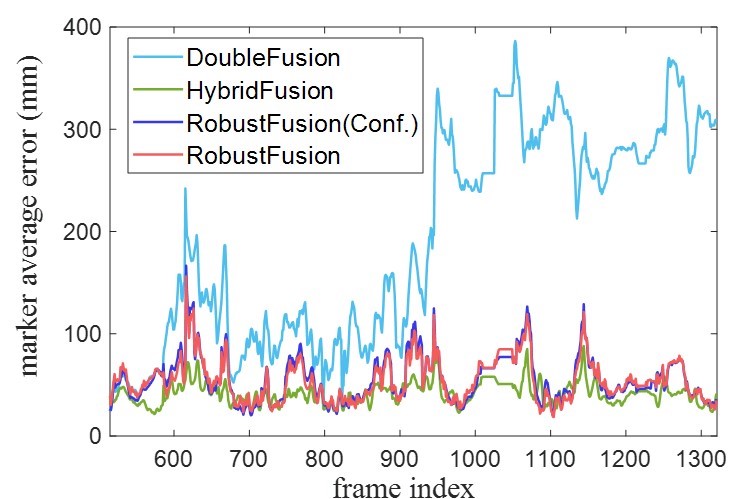}}
	\caption{Quantitative comparison. (a-d) are the reconstruction geometry results of DoubleFusion~\cite{DoubleFusion}, HybridFusion~\cite{HybridFusion}, RobustFusion(Conf.)~\cite{RobustFusion2020ECCV} and our method. (e) is the error curves.}
	\label{fig_compare_3}
\end{figure}

To illustrate our robustness for human-specific motions, we further compare against HybridFusion\cite{HybridFusion}, which uses extra body-worn IMU sensors.
We utilize the challenging sequence with ground truth from \cite{HybridFusion} and remove their orchestrated self-scanning process for our methods.
Even though this sequence does not include human-object interaction scenarios, such an experiment further illustrates that our approach with data-driven interaction cues can handle challenging human-only motions.
The quantitative comparison in terms of the per-frame error in Fig.~\ref{fig_compare_3} (e) and the average errors among the whole sequence in Tab.\ref{table:comp2} demonstrate that both our approach and our preliminary version~\cite{RobustFusion2020ECCV} achieve a significantly better result than DoubleFusion and even comparable performance against HybridFusion.
Note that HybridFusion still relies on the self-scanning stage for sensor calibration and suffers from missing geometry caused by the body-worn IMUs as shown in Fig.~\ref{fig_compare_3} (a), while our approach eliminates such tedious self-scanning and achieves complete and plausible reconstruction results.  

\begin{table}[htbp]
	\footnotesize
	\newcommand{\tabincell}[2]{\begin{tabular}{@{}#1@{}}#2\end{tabular}}
	\renewcommand{\arraystretch}{1.3}
	\addtolength{\tabcolsep}{-2.5pt}
	\centering
	\caption{Average errors on the entire sequence compared to the ground truth observation from the Vicon system, for these three methods: DoubleFusion ~\cite{DoubleFusion}, HybridFusion~\cite{HybridFusion}, RobustFusion(Conf.)~\cite{RobustFusion2020ECCV} and our method, respectively.}
	\begin{tabular}{p{1.8 cm}p{1.3cm}p{1.3cm}p{1.3cm}p{1.3cm}}
		\hline
		& \tabincell{c}{~\cite{DoubleFusion}} & \tabincell{c}{~\cite{HybridFusion}}  & \tabincell{c}{~\cite{RobustFusion2020ECCV}} & \tabincell{c}{Ours}\\
		\hline	    
		{\small{\textit{average error}}}& 0.1904 m  & 0.0417 m & 0.0553 m & 0.0546 m\\		
		\hline
	\end{tabular}
	\label{table:comp2}
\end{table}

\subsection{Evaluation}\label{Sec:evaluation}
In this subsection, we evaluate each technical contribution of our RobustFusion separately. 
Specifically, we evaluate the human initialization, mask refinement, object tracking, robust human tracking, and object-aware adaptive fusion, respectively.
Moreover, we also evaluate our extension capability by experiments in multi-person and multi-camera scenarios. 

\begin{figure}[htb]
	\centering
	\subfigure[]{\includegraphics[height=220pt]{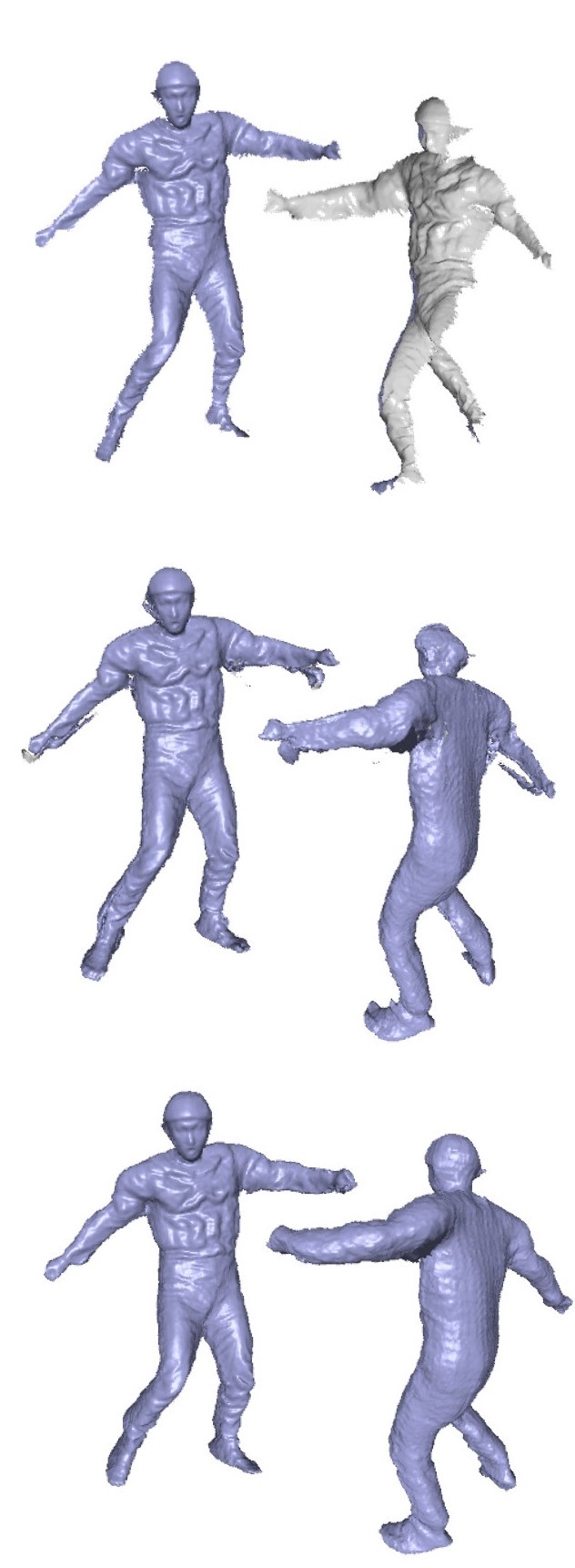}\label{fig:init_a}}
	\subfigure[]{\includegraphics[height=220pt]{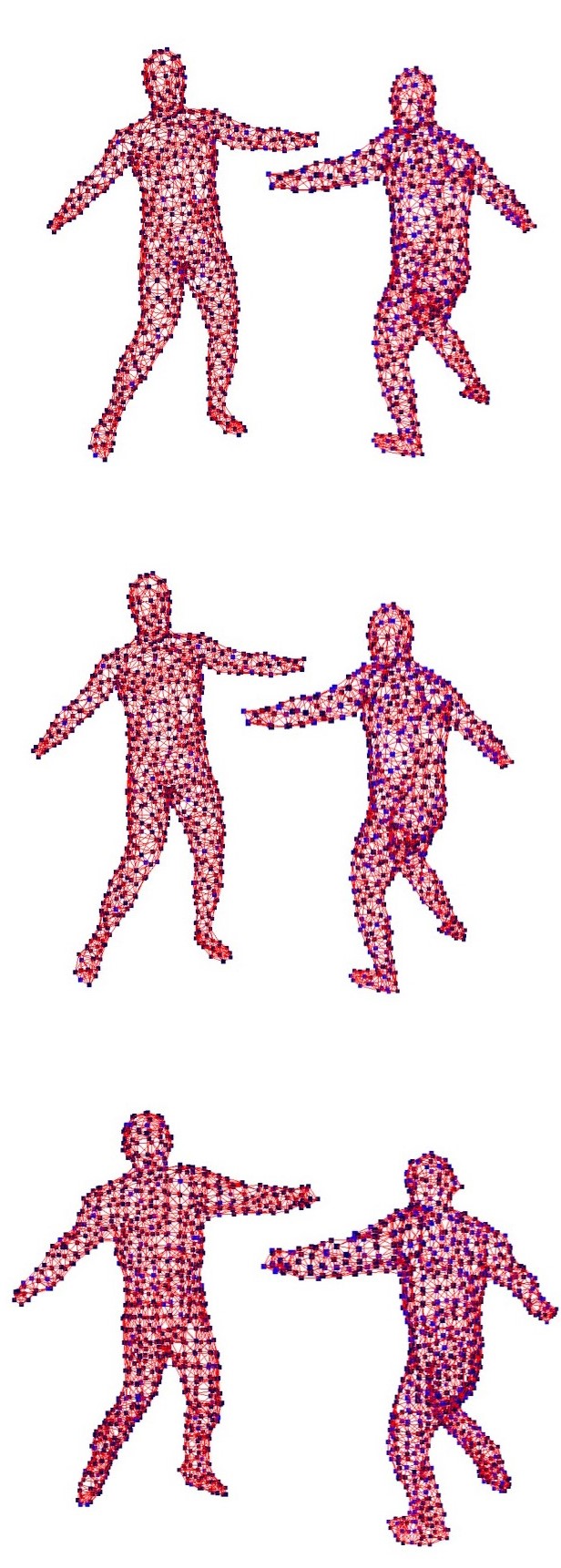}\label{fig:init_b}}	\subfigure[]{\includegraphics[height=220pt]{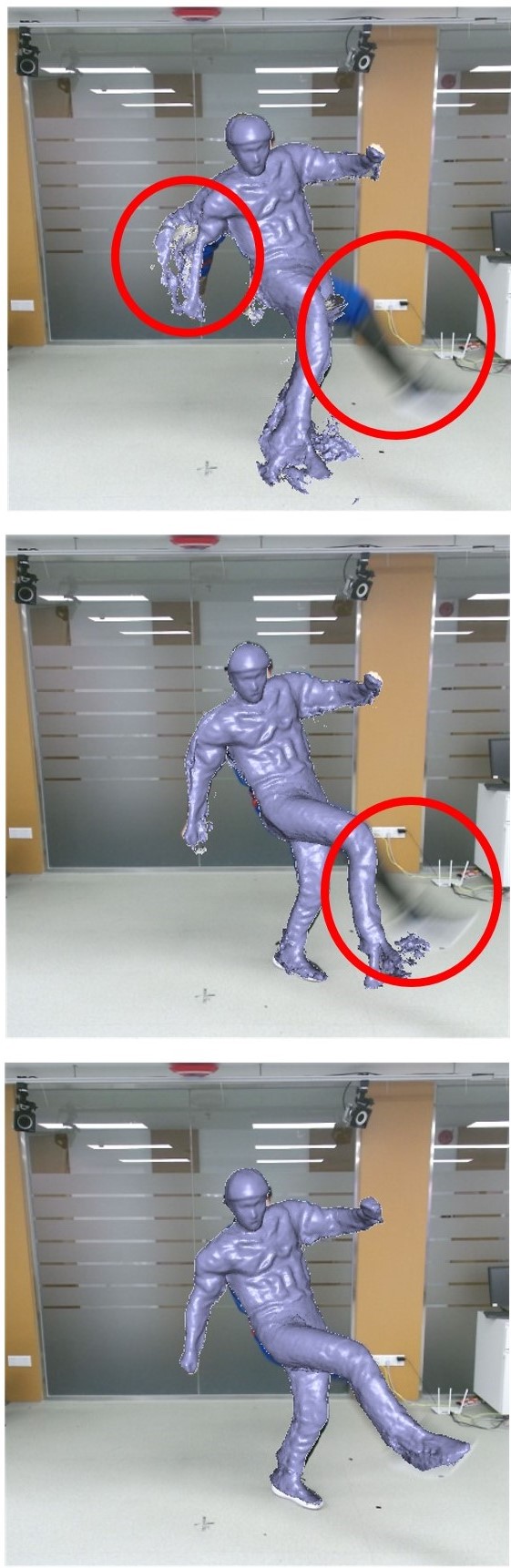}\label{fig:init_c}}
	\caption{Evaluation of human initialization. (a) is the 3D human model in two views. (b) is the ED node-graph that formulates the surface motions. (c) is the following tracking results that overlay on reference color image based on the corresponding 3D surface geometry and the ED node-graph.The results from the first row to the third are the results without human initialization, with initialization only using skeleton optimization and SMPL-based node-graph, and with our entire initialization process, respectively.}
	\label{fig_ablation_1}
\end{figure}

\noindent\textbf{Human Initialization.}
For completeness of evaluation, we first evaluate the human initialization scheme on a sequence without a carefully designed self-scanning process organized as model completion and initialization in performance capture stages in~\cite{RobustFusion2020ECCV}.  
As shown in Fig.~\ref{fig_ablation_1}, without model initialization, only partial initial geometry with SMPL-based ED node-graph leads to inferior tracking and erroneous reconstruction results. 
This exactly explains that why DoubleFusion~\cite{DoubleFusion} and UnstructuredFusion~\cite{UnstructureLan} fail without careful self-scanning process.
To evaluate our alignment during model initialization and demonstrate the superiority of our modified motion representation over original representation in previous methods~\cite{DoubleFusion,UnstructureLan}, the skeletal pose is optimized during alignment optimization, and only SMPL-based double-layer ED-graph is adopted for motion tracking, where the results are still imperfect.
In contrast, our approach with both model and motion initialization successfully obtains a watertight and fine-detailed human mesh and enables more robust motion tracking.

\begin{figure}[htb]
	\centering
	\subfigure[]{\includegraphics[height=150pt]{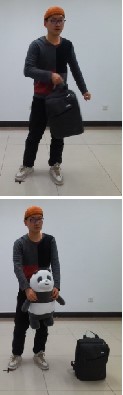}\label{fig:mask_a}}
	\subfigure[]{\includegraphics[height=150pt]{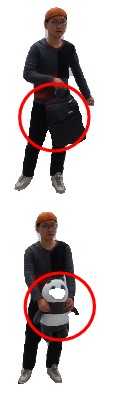}\label{fig:mask_b}}	\subfigure[]{\includegraphics[height=150pt]{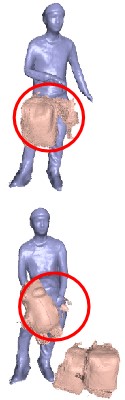}\label{fig:mask_c}}
	\subfigure[]{\includegraphics[height=150pt]{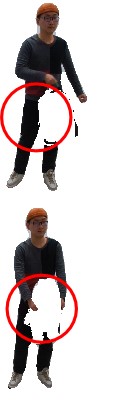}\label{fig:mask_d}}	\subfigure[]{\includegraphics[height=150pt]{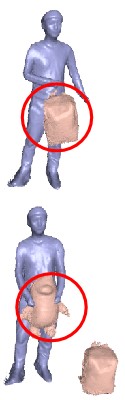}\label{fig:mask_e}}
	\caption{Evaluation of the mask refinement. (a) Reference color images. (b) The human masks without refinement. (c) The reconstruction results without refinement. (d) The human masks with refinement. (e) The reconstruction results with refinement.}
	\label{fig_eva_mask}
\end{figure}

\begin{figure}[htb]
	\centering
	\includegraphics[width=1.0\linewidth]{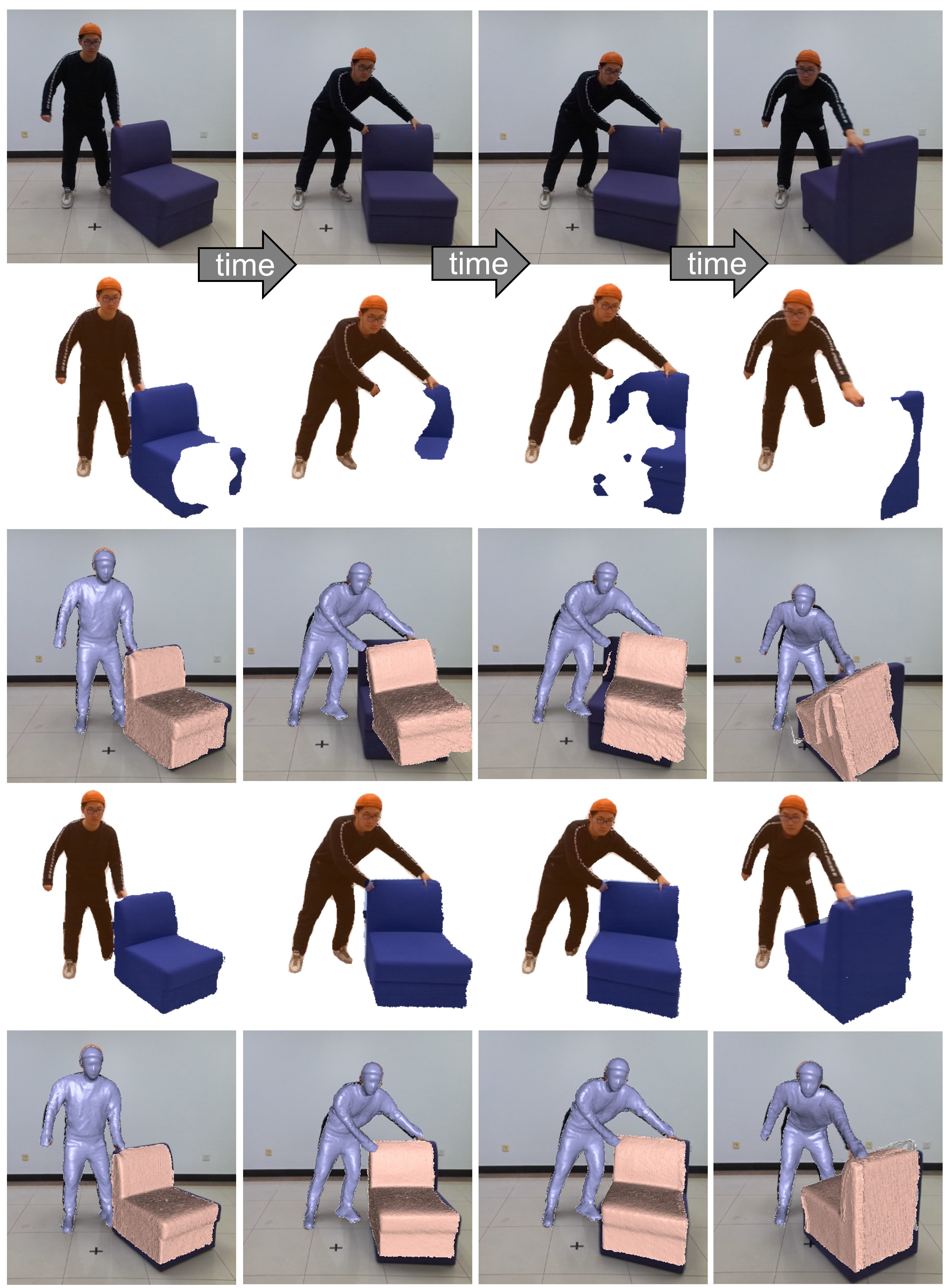}
	\caption{Evaluation of the mask refinement (for objects). The first row demonstrates the input color images. The second and the fourth row are the original masks from the network~\cite{zhao2017pspnet} and our refined masks, respectively. The third and the last row overlaying on the color images are the tracking results using the object masks before and after mask refinement, respectively.}
	\label{fig_eva_mask_2}
\end{figure}

\noindent\textbf{Mask Refinement and Object Tracking.}
Here we evaluate the proposed mask refinement in Fig.~\ref{fig_eva_mask}.
Since the toy is unlabeled in the network~\cite{zhao2017pspnet}, we extract its masks by utilizing background subtracting. 
Therefore, the segmentation of objects is also dependent on human segmentation results. 
The original human segmentation mask in Fig.~\ref{fig_eva_mask} (b) is inaccurate, especially when human-object interaction occurs, leading to misaligned object masks due to the overlaying of the object on the performer. 
Then, directly using such coarse segmentation results leads to unstable tracking and erroneous object reconstructions as shown in Fig.~\ref{fig_eva_mask} (c).
In contrast, with our mask refinement scheme, the object is separated from the human segmentation result correctly (Fig.~\ref{fig_eva_mask} (d)). 
As a result, our approach achieves more robust human and object tracking results in Fig.~\ref{fig_eva_mask} (e), which illustrates the effectiveness of our layer-wise strategy and mask refinement scheme.

\begin{figure}[htb]
	\centering
	\includegraphics[width=0.90\linewidth]{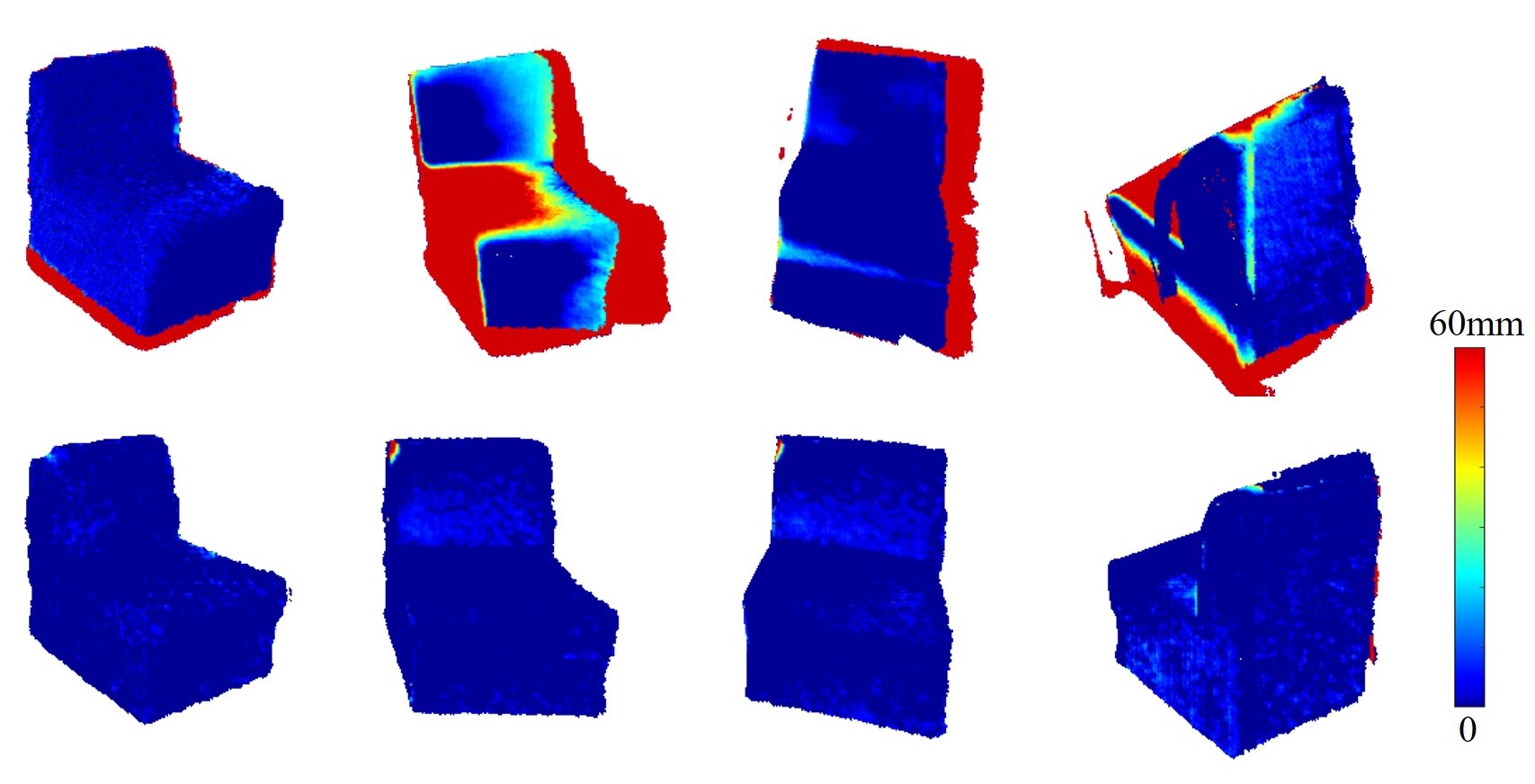}
	\caption{Evaluation of the mask refinement (for objects). The color-coded maps indicate the projective error maps, in which the two rows are corresponding to the results in Fig.~\ref{fig_eva_mask_2}.}
	\label{fig_eva_mask_3}
\end{figure}

Besides, Fig.~\ref{fig_eva_mask_2} further demonstrates the robustness of our mask refinement for object mask segmentation and rigid tracking. 
Although sofa/chair is labeled in the network~\cite{zhao2017pspnet}, it occasionally fails to extract the masks as shown in the second row of Fig.~\ref{fig_eva_mask_2}. 
With the refined masks and object tracking optimization in the fourth row of Fig.~\ref{fig_eva_mask_2}, we can achieve more accurate object tracking. 
The corresponding quantitative comparison in Fig.~\ref{fig_eva_mask_3} further demonstrates that our method achieves the highest accuracy, where the MAE for the entire object sequence is 35.10 mm and 11.11 mm for our method w/o and with mask refinement, respectively.

\begin{figure}[htb]
	\centering
	\subfigure[]{\includegraphics[width=0.32\linewidth]{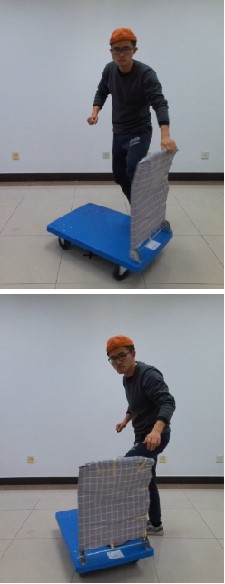}}
	\subfigure[]{\includegraphics[width=0.32\linewidth]{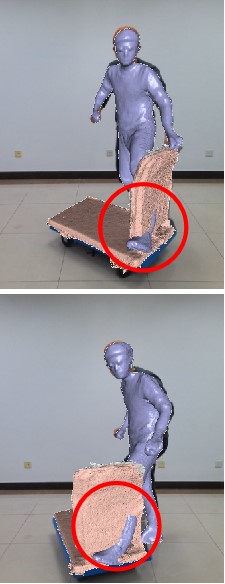}}	
	\subfigure[]{\includegraphics[width=0.32\linewidth]{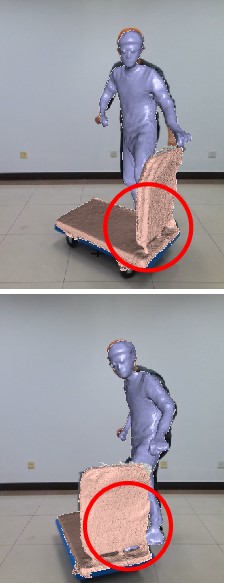}}
	\caption{Evaluation of the interpenetration term. (a) Reference color images. (b) The results without interpenetration term. (c) The results with interpenetration term.}
	\label{fig_eva_space}
\end{figure}

\begin{figure}[t]
	\centering
	\subfigure[Fast motion scenarios]{\includegraphics[width=0.95\linewidth]{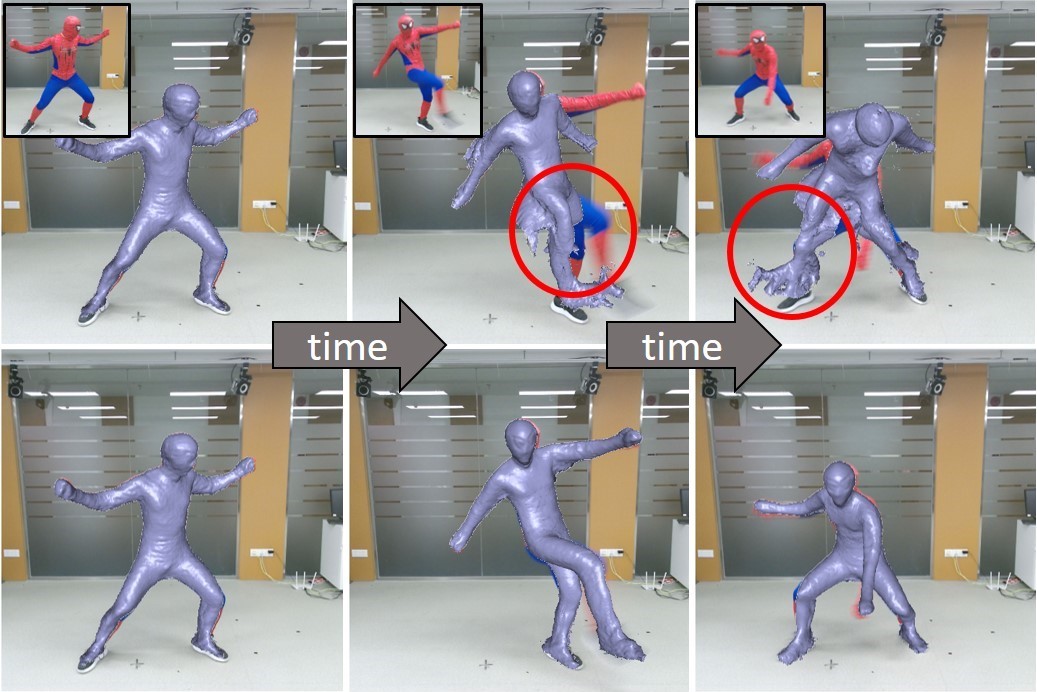}}
	\subfigure[Disappear and reoccur scenarios]{\includegraphics[width=0.95\linewidth]{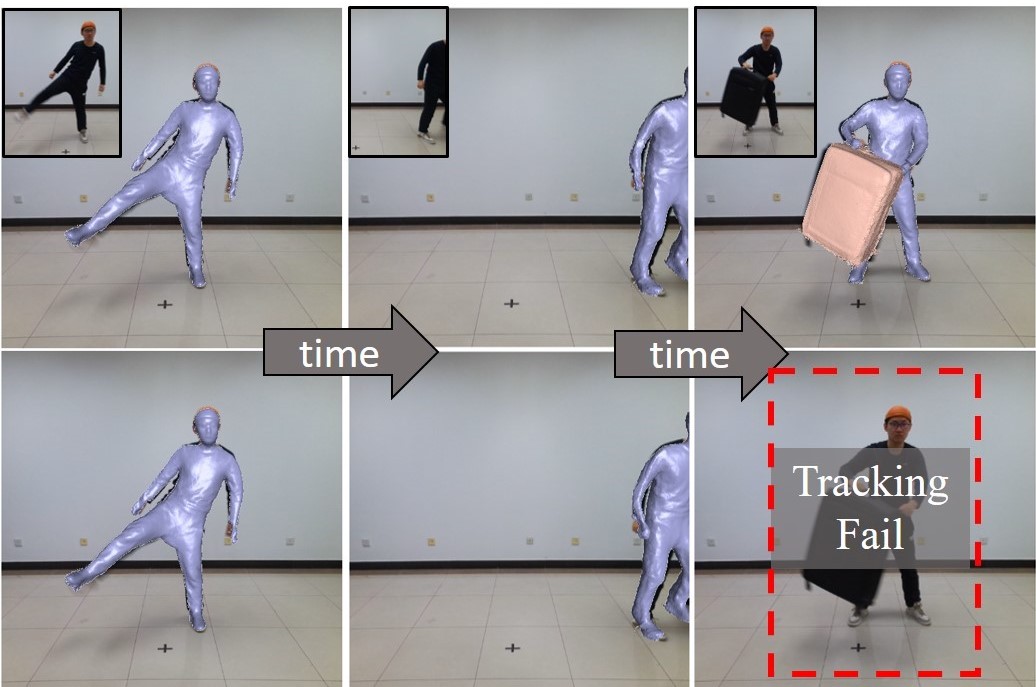}}	
	\caption{Evaluation of pose term. The top row of (a) and (b) are results of our method variation without pose term. The bottom row of (a) and (b) are results of our approach with pose term in optimization.
	}
	\label{fig_eva_pose}
\end{figure}

\noindent\textbf{Robust Human Tracking.}
Our robust human tracking is based on human-object spatial relation analysis using various data-driven cues. Here, we evaluate them one by one.
First, as shown in Fig.~\ref{fig_eva_space}, we evaluate our human-object spatial relation cue -- the interpenetration term for human motion optimization.
Note that we eliminate the interpenetration term by setting $\lambda_{sp\_h1} = 0$ and $\lambda_{sp\_h2} = 0$ in Fig.~\ref{fig_eva_space} (b), where the human model erroneously inserts into the cart.
Differently, our full pipeline provides an essential spatial constraint for human motions estimation, especially in occlusion cases where no direct observation is available like the leg in Fig.~\ref{fig_eva_space}. 
Benefit from our interpenetration term, we successfully avoid the interpenetration between human and cart models as demonstrated in Fig.~\ref{fig_eva_space} (c).

Then, we evaluate the data-driven visual cues -- the pose term in human motion optimization. 
Similar to the preliminary method~\cite{RobustFusion2020ECCV}, we compare to the variation of our pipeline without pose prior in two scenarios where fast motion or disappear-reoccurred case happens. 
The first row of Fig.~\ref{fig_eva_pose} demonstrates that our variation without pose term suffers from severe accumulated error, especially for the limb region with faster motions and more depth noise from the commercial sensor. 
Our full pipeline relieves this problem and help achieve superior tracking results for these challenging cases.
Besides, as shown in the second row of Fig.~\ref{fig_eva_pose}, with the aid of pose detection and scene segmentation, the system can screen the disappears and re-occur of the person with reconstructing object and achieve the recovery from the failing track.

\begin{figure}[htb]
	\centering
	\subfigure[]{\includegraphics[height=140pt]{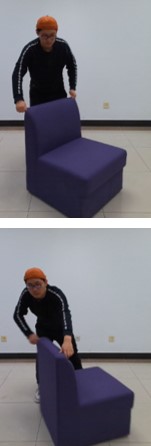}\label{fig:prior_a}}
	\subfigure[]{\includegraphics[height=140pt]{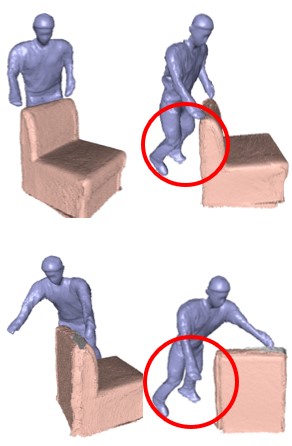}\label{fig:prior_b}}
	\subfigure[]{\includegraphics[height=140pt]{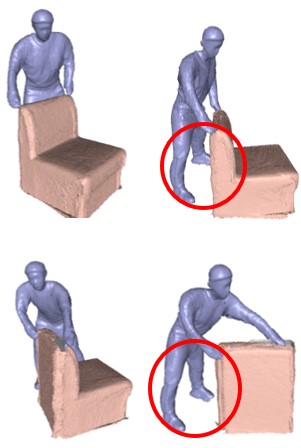}\label{fig:prior_c}}
	
	\caption{Evaluation of the prior constraints. (a) Reference color images. (b) The results without prior constraints (front view and side view respectively). (c) The results with prior constraints (front view and side view respectively).}
	\label{fig_eva_prior}
\end{figure}

Furthermore, the evaluation of empirical data-driven terms, including an interaction pose prior and a temporal motion prediction prior, is provided in Fig.~\ref{fig_eva_prior}. 
We eliminate the data-driven interaction terms by setting $\lambda_{lstm} = 0$ and $\lambda_{gmm} = 0$.
Then, due to the severe occlusions between the performer and the sofa, this variation generates unnatural motion estimation for the occluded legs as shown in the different rendered views in Fig.~\ref{fig_eva_prior} (b). 
With the aid of the empirical constraint, our approach can generate more plausible and reasonable results, as shown in the corresponding sub-figures (c) of Fig.~\ref{fig_eva_prior}. 
In this comparison, the pose estimation of the unobserved region is up to the continuity of observed joints in the kinematic chain, similar to the unconstrained optimization that may lead to severe deviation from the real situation. 
Differently, the human skeletal pose estimation with an empirical constraint can be well deducted by the historical experience, including both interaction pose distribution prior based on Gauss Mixture Model and temporal motion prediction based on LSTM.

\begin{figure}[htb]
	\centering
	\subfigure[]{\includegraphics[height=190pt]{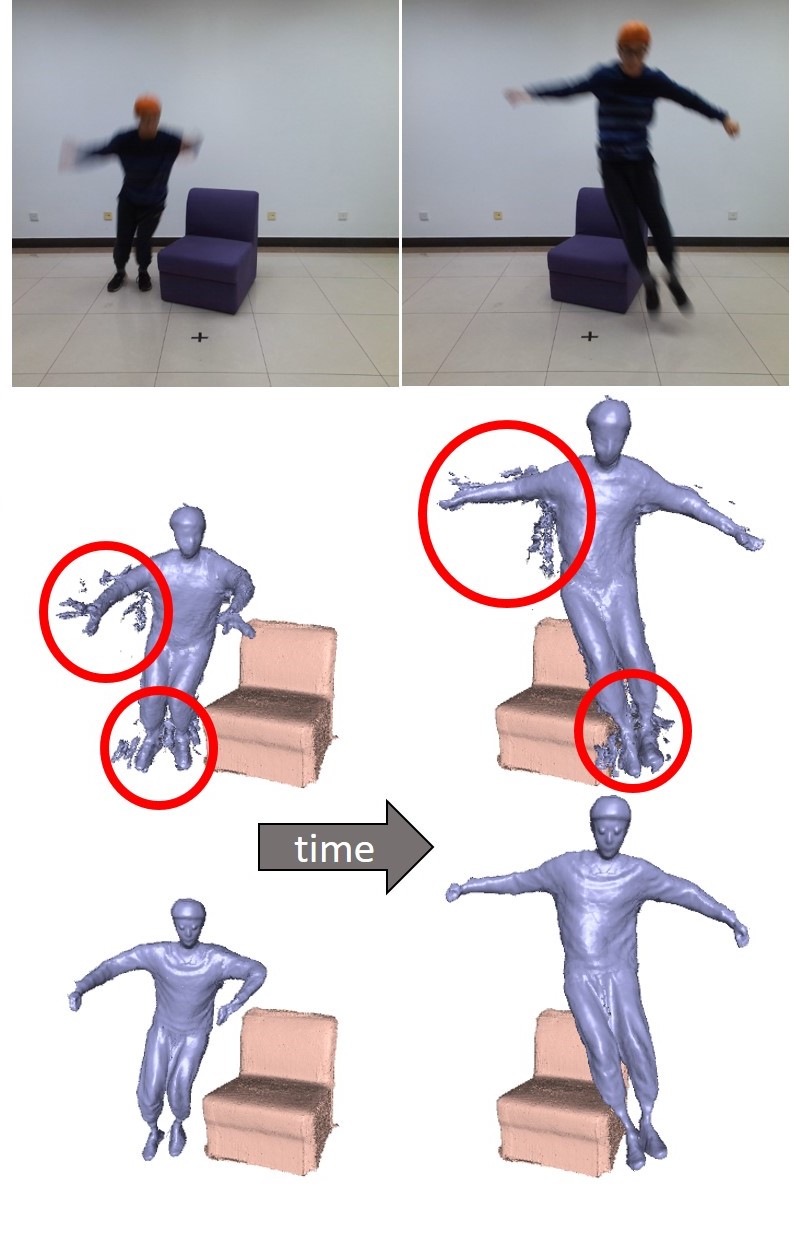}\label{fig:fusion_a}}
	\subfigure[]{\includegraphics[height=190pt]{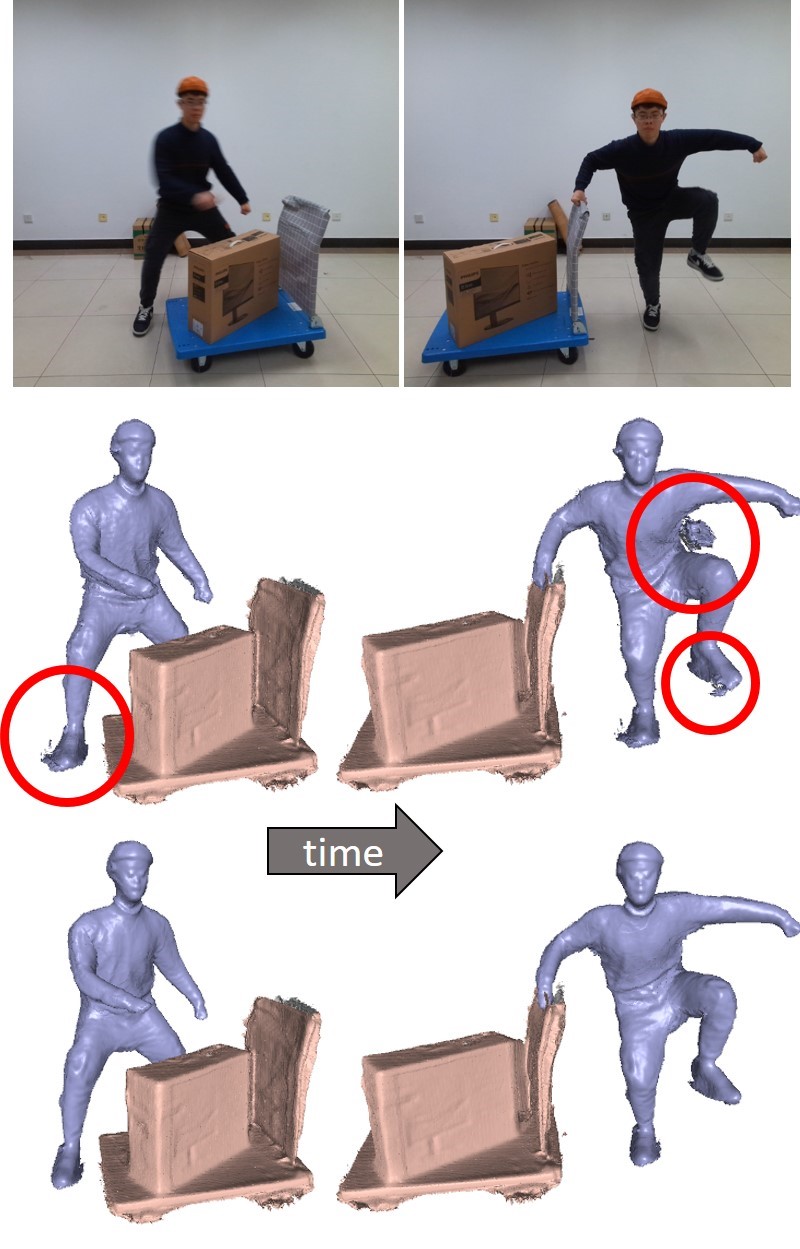}\label{fig:fusion_b}}
	\caption{Evaluation of the object-aware adaptive fusion. (a) A sequence with high-speed motions. (b) A sequence with occlusions. The first to the third row are reference color images, the geometry results without object-aware adaptive fusion, and the geometry results with object-aware adaptive fusion.}
	\label{fig_eve_fusion}
\end{figure}

\noindent\textbf{Object-aware Adaptive Fusion.}
To evaluate our object-aware adaptive fusion scheme based on the occlusion relation and semantic errors, we compare our full volumetric fusion pipeline and the method variation without the object-aware adaptive fusion strategy.
The comparison in Fig.~\ref{fig_eve_fusion} (a) demonstrates that our full pipeline can effectively avoid the severe accumulated error for those regions with high-speed motions, such as jumping over the chair.
Besides, as shown in the second row of Fig.~\ref{fig_eve_fusion} (b), the occlusion by object leads to erroneous surface generation (e.g., when the performer pulls the cart and walks behind the object). 
In contrast, the geometry results in the bottom row of Fig.~\ref{fig_eve_fusion} (b) demonstrate that our object-aware adaptive fusion can successfully model occluded scenarios and avoid deteriorated fusion.

\noindent\textbf{Expansion for Multi-person scenarios.}
Here we show our capability for extending to multi-person capture.
As demonstrated in the last row of Fig.~\ref{fig_all_result}, with semantic segmentation labels of different subjects of the whole scene and our mask refinement process, we can also enable multi-person reconstruction. 
Note that to capture such a larger scene with two persons, we deploy a two-camera system using the online calibration from \cite{UnstructureLan}. 
We believe that it is an essential step for the reconstruction of more general dynamic scenes.

\subsection{Limitation}
As a trial for robust monocular volumetric performance capture under human-object interactions, we have demonstrated compelling 4D reconstruction results. Nonetheless, our approach is subject to some limitations.

Similar to the previous methods~\cite{DoubleFusion,UnstructureLan,li2021posefusion,RobustFusion2020ECCV}, our method cannot reconstruct the extremely fine details of the performer, such as the fingers, the subtle expression, and shaggy hair, due to the limited resolution and inherent noise of the depth input. 
It is promising to adopt data-driven techniques to further generate visually pleasant synthetic geometry details in those model-specific regions. 
Besides, the reconstruction of loose and wide cloth such as a long skirt with high-speed motions remains challenging since it is difficult to track such large free-form non-rigid deformation beyond human skeletal motions. 
It is also challenging for human initialization in Sec.~\ref{Sec:initialization}. 
A better human model regression algorithm during model initialization or utilizing a pre-scanned detailed template will help remove this limitation.
Furthermore, we cannot handle surface splitting topology changes like clothes removal, which we plan to address by incorporating the key-volume update technique~\cite{dou-siggraph2016}.
As common for learning methods, the utilized scene semantic segmentation, human parsing, and pose estimation fail for extreme scenarios not seen in training, such as severe and extensive (self-)occlusions under extreme side-view observation. 
However, our mask refinement strategy turns to obtain accurate masks, and data-driven cues of motion prediction and pose prior help us to relieve the occlusion problem with re-initialization ability.
As for more general interactions, our current system still cannot handle tiny objects which can be played with fingers or non-rigid objects like dolls or papers, which restricts the wide practical applications of our approach.
The limitation of tiny objects is also due to the limited image resolution and quality of the available commercial RGBD sensors.
We plan to combine those the task-specific approach such as~\cite{GrapingField:3DV:2020,GRAB:2020} for fine-grained interaction modeling and extend our method to non-rigid objects by modifying our non-graph sampling and updating strategy.
Besides, our current pipeline has tried to handle multi-person scenarios with inter-person interactions at a certain level by using more RGBD sensors as input.
It is a promising and challenging direction to deal with more general inter-person interactions such as dancing, wrestling, and hugging, even using the same monocular RGBD input.

	\section{Conclusion}
We have presented RobustFusion, a robust volumetric performance reconstruction approach for complex human-object interactions and challenging human motions using only a single RGBD sensor.
It combines various data-driven visual and interaction cues for robust human-object 4D reconstruction whilst still maintaining light-weight computation and monocular setup.
Our scene decoupling scheme with segmentation refinement and robust object tracking enables explicit human-object disentanglement and temporal-consistent modeling, while our human initialization gets rid of the tedious self-scanning constraint.
Our robust human performance capture with various visual and interaction cues models complex interaction patterns in a data-driven manner and enables natural motion reconstruction under challenging human-object occlusions, with unique re-initialization ability.
Our object-aware adaptive fusion scheme successfully avoids deteriorated fusion and obtains temporally coherent human-object reconstruction with the aid of occlusion analysis and human parsing cue.
Extensive experimental results demonstrate the effectiveness and robustness of our approach for compelling performance capture in various challenging scenarios with human-object interactions.
We believe that it is a critical step to enable robust and lightweight dynamic scene reconstruction under human-object interactions, with many potential applications in VR/AR, entertainment, human behavior analysis, and immersive telepresence.

	\appendices

	\ifCLASSOPTIONcaptionsoff
	\newpage
	\fi

	{\small
		\bibliographystyle{IEEEtran}
		\bibliography{RobustFusionPlus}
	}

\end{document}